\documentclass{article} % For LaTeX2e
\usepackage{colm2024_conference}

\usepackage{microtype}
\usepackage{hyperref}
\usepackage{url}
\usepackage{booktabs}
\definecolor{darkblue}{rgb}{0, 0, 0.5}
\hypersetup{colorlinks=true, citecolor=darkblue, linkcolor=darkblue, urlcolor=darkblue}

\usepackage{amsmath}
\usepackage{svg}
\usepackage{amssymb}
\usepackage{algorithm}
\usepackage{algpseudocode}
\usepackage{colortbl}
\usepackage{caption}
\usepackage{subcaption}
\usepackage{multicol}
\usepackage{wrapfig}
\usepackage{blindtext}
\usepackage{multirow}

\title{VTrans: Accelerating Transformer Compression with Variational Information Bottleneck based Pruning}

\author{
Oshin Dutta \textsuperscript{1} \quad Ritvik Gupta \textsuperscript{2} \quad Sumeet Agarwal \textsuperscript{1}\\
\textsuperscript{1} Indian Institute of Technology, Delhi \\
\textsuperscript{2} Carnegie Mellon University \\
\href{mailto:oshin.dutta@ee.iitd.ac.in}{oshin.dutta@ee.iitd.ac.in}, \href{mailto:ritvikgu@cs.cmu.edu}{ritvikgu@cs.cmu.edu}, \href{mailto:sumeet@iitd.ac.in}{sumeet@iitd.ac.in}
}

\colmfinalcopy % Uncomment for camera-ready version, but NOT for submission.
\begin{document}

\maketitle

\begin{abstract}
In recent years, there has been a growing emphasis on compressing large pre-trained transformer models for resource-constrained devices. However, traditional pruning methods often leave the embedding layer untouched, leading to model over-parameterization. Additionally, they require extensive compression time with large datasets to maintain performance in pruned models.
To address these challenges, we propose \textbf{VTrans}, an iterative pruning framework guided by the Variational Information Bottleneck (VIB) principle. Our method compresses all structural components, including embeddings, attention heads, and layers using VIB-trained masks. This approach retains only essential weights in each layer, ensuring compliance with specified model size or computational constraints. Notably, our method achieves upto 70\% more compression than prior state-of-the-art approaches, both task-agnostic and task-specific. We further propose faster variants of our method: \textbf{Fast-VTrans} utilizing only 3\% of the data and \textbf{Faster-VTrans}, a time efficient alternative that involves exclusive finetuning of VIB masks, accelerating compression by upto 25 times with minimal performance loss compared to previous methods. Extensive experiments on BERT, ROBERTa, and GPT-2 models substantiate the efficacy of our method. Moreover, our method demonstrates scalability in compressing large models such as LLaMA-2-7B, achieving superior performance compared to previous pruning methods. Additionally, we use attention-based probing to qualitatively assess model redundancy and interpret the efficiency of our approach.
Notably, our method considers heads with high attention to special and current tokens in un-pruned model as foremost candidates for pruning while retained heads are observed to attend more to task-critical keywords.
\end{abstract}
\section{Introduction}
%%Why do we want to compress transformer architecture- specially BERT
% The need for compressed models to deploy on edge device for fast inference \\
Since their inception, Transformers \citep{transformer2017} have fundamentally transformed the NLP field, offering pre-trained self-supervised models adaptable to specific downstream tasks \citep{wang-etal-2018-glue}. However, the surge in popularity and scale of Transformer models has amplified deployment challenges on resource-constrained devices, attributed to elevated latency and substantial storage demands \citep{raffel2020exploring,brown2020language}.

This has motivated extensive work in transformer pruning focusing on various components such as layers, heads, blocks within weight matrices, and hidden states. However, embeddings, which account for over 22\% of total model parameters, are often overlooked \citep{cofi2022,dynabert} due to challenges in maintaining consistency with skip connection. Moreover, existing methods often use magnitude-based~\citep{han_et_al2015} or simply sparsity-based pruning~\citep{cofi2022}, which overlooks the importance of weights for the given task.
%Moreover, existing methods often enforce uniform hidden state sizes across all transformer feedforward layers. %This overlooks the fact that different layers capture varying aspects of the data, ultimately limiting the potential benefits of pruning. 

An alternative approach to achieving compact yet high-performing models is knowledge distillation \citep{hinton2015distilling}, where insights from a larger teacher model are transferred to a carefully designed smaller student model. However, task-agnostic knowledge distillation can incur a prohibitively high computational cost \citep{jiao2019tinybert}. Quantization is another effective approach that achieves high compression and speedup with minimal loss in performance by leveraging techniques such as quantization-aware training \citep{bai2021binarybert, kim2021ibert, q8bert_Zafrir_2019}.

%This has motivated the extensive work in model compression of pre-trained models, mainly in pruning and distillation methods. Distillation methods define a more compact architecture for the student model, followed by the application of knowledge distillation \citep{hinton2015distilling} on the student model using unlabelled data and a teacher model. In case of pruning methods, the idea is to identify redundancy in the structural elements and prune such elements to obtain a smaller sub-network from a larger model.
\begin{figure*}[t]
\centerline{\includesvg[inkscapelatex=false,width=\textwidth]{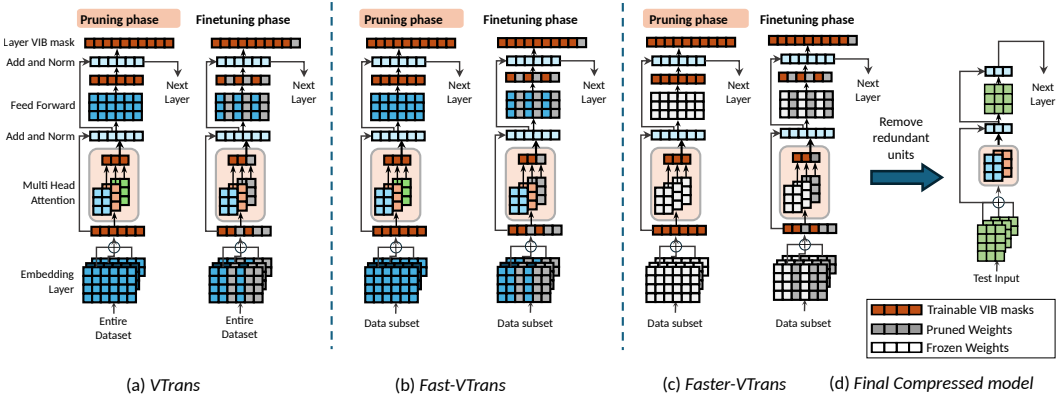}}   
\caption{(a) Our primary method - \textit{VTrans} involves training all pre-trained model parameters and VIB masks on the entire dataset during pruning, but during finetuning, only the unmasked important weights are updated.
(b) \textit{Fast-VTrans} utilizes a subset of data for both pruning and finetuning.
(c) \textit{Faster-VTrans}, the fastest among all, trains VIB masks, Add Norm layer, and model bias parameters during pruning and finetuning, using only a subset of the entire dataset.
(d) The masked and redundant units are removed after the finetuning phase, resulting in the dense compressed model.}
    \label{fig:1}   
\end{figure*}
To address the challenges, we propose a novel transformer pruning method based on the Variational Information Bottleneck (VIB) principle. It effectively removes redundant elements while preserving information flow. Additionally, we leverage knowledge distillation to achieve higher pruning ratios without sacrificing performance. 

\begin{wrapfigure}[17]{t}{0.5\textwidth}
\includesvg[inkscapelatex=false,width=0.5\textwidth]{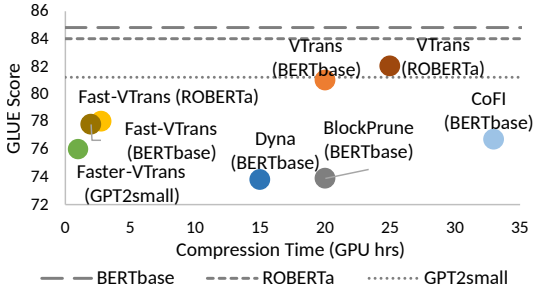}
    %BERT_pic_1_wider.svg
%\includesvg[inkscapelatex=false,width=\textwidth]{images/gpt2_maskonly_speedup.svg}
\caption{Our method surpasses previous techniques in compressing BERT-base, ROBERTa-base, and GPT-2-small models. Our faster variants have reduced compression timeframes. All models have 28M parameters being compressed from their respective teachers.}\label{fig:overall}
 \end{wrapfigure}

Our contributions are:
\begin{itemize}
    \item We propose a structured pruning framework for transformers that is evaluated in both task-specific and task-agnostic contexts, while adhering to user-defined constraints on either model parameters or FLOPs.
    \item Unlike prior methods, we extend compression to embedding states, alongside other structural components like layers, attention heads, and feedforward networks (FFN) enabling higher compression levels.% by adeptly adjusting layer sizes to seamlessly integrate with the compressed embeddings. 
    %This leads to a configuration that is highly adaptable and well-suited for deployment on devices with limited resources.
    %\item Our approach, tested on BERT \citep{devlin-etal-2019-bert} achieves an additional compression of 20\% to 70\% compared to previous methods and upto 60\% reduction in FLOPs in DistilBERT~\citep{sanh2020distilbert} while maintaining performance within 1\% of the base model on both GLUE and SQuAD tasks.
    \item We propose two alternative - \textit{Fast} and \textit{Faster} approaches (Figure~\ref{fig:1}) to the main compression method which are both time and resource efficient with minimal performance degradation compared to previous approaches.
    \item We evaluate the proposed methods on GLUE and SQuAD tasks while compressing BERT, ROBERTa and GPT-2 pre-trained models with superior performance to previous SOTA (Figure~\ref{fig:overall}).
    \item Furthermore, we establish the scalability of our method by pruning and evaluating LLaMA-2 with 7 billion parameters on WikiText-2 dataset (Table~\ref{tab:llama_prune}).
    %show preservation of grammatical nuances and reduction of correlation in the attention heads in the compact model.
    %\item We mathematically prove reducation of correlated structures
    %\item We provide qualitative results with an in-depth analysis of attention heads, to reveal the the impact of pruning and visually demonstrate redundancies in the BERT architecture.
    %give some numbers
    %\item We deploy the compressed models on popular resource restricted devices and compare the inference speed and discuss about the ease of deployment. 
\end{itemize}
%%% Pruning in BERT;  pruning all possible structural elements and knowledge Distillation
% Need for a principled pruning approach that prunes redundant structural elements in BERT while keeping the information flow of the model intact. \\
% Use of knowledge distillation for higher pruning ratios while preserving performance.\\
%%what Do we propose
%%Read intro from train-flat and then compress paper.
%~\citep{kwon_fast_2022, yang_sparse_2022, na_train_2022,liang_homodistil_2023, kurtic_ziplm_2023}
%cite- A Fast Post-Training Pruning Framework for Transformers- uses flops constraints
%compare results with- Train Flat, Then Compress:Sharpness-Aware Minimization Learns More Compressible Models
\section{Related Work}
\textbf{Pruning.} It involves removing redundant model parameters for substantial model 
compression with minimal performance loss \citep{cofi2022,zhu2017prune,zafrir2021prune,renda2020comparing,liang2023less}. Recent techniques for transformer model pruning employ structured approaches, targeting specific components like layers \citep{Fan2020layerdrop,Sajjad2020PoorMB}, heads \citep{michel_heads,voita_analyzing_2019,liang2021super}, intermediate dimensions \citep{mccarley2021structured,wang_structured_2020} and blocks of weight matrices \citep{lagunas-etal-2021-block} or multiple components jointly \citep{cofi2022,sun2023simple}. Low-rank approximation has also been combined with pruning \citep{li2023losparse} to further compression. But these prior methods do not prune embedding parameters which often form more than 20\% of the total parameters. 
%Additionally, they often overlook the importance of weights for the given task when pruning. 
Methods utilizing only forward pass during pruning~\citep{sun2023simple} often involve semi-structured pruning that cannot be made dense post-pruning, leading to slower inference speeds. Certain pruning methods focus only on task-agnostic~\cite{liang_homodistil_2023} and some on task-specific~\cite{nasery2023end,yang2022task}. %and do not support flexible layer sizes in models as consistent dimensions are essential in skip connections.

% All prior pruning methods utilize gradient-free compression techniques, followed by fine-tuning of the resulting model.
% \begin{wraptable}[11]{t}{0.45\textwidth}
% \resizebox{0.45\textwidth}{!}{
% \begin{tabular}{lccc}
% \hline
% Model & Size & PPL & Inference Speedup  \\
% \hline
% LLaMA-2 (teacher) & 7B & 5.11 & 1x \\
% Phi-2 & 3B & 8.69 & 1.24× \\
% Wanda 2:4 &3B & 8.34 & 0.75× \\
% Bonsai & 3B & 8.89 &1.58× \\
% Bonsai w/o FT & 3B & 19.47 &1.58× \\
% \hline
% \rowcolor[gray]{0.8}
% \textbf{Faster-VTrans w FT}  & 3B & 8.62 & 1.76x\\
% \rowcolor[gray]{0.8}
% \textbf{Faster-VTrans w/o FT}  & 3B &  14.19 & 1.76x\\
% \hline
% \end{tabular}}
% \captionof{table}{Performance of models pruned by 50\% from LLaMA-2-7B evaluated on Wikitext-2.}\label{tab:llama}
% \end{wraptable}
\textbf{Knowledge distillation.} Another compression method involves training smaller (student) model from a larger (teacher) model. %The student model aims to match the outputs of the teacher model and is penalized for any deviations from the teacher's outputs. 
The technique has found applications in both, task-specific \citep{tang2019distilling,turc2019wellread,aguilar2020knowledge} as well as task-agnostic domains \citep{sanh2020distilbert,khanuja-etal-2021-mergedistill,chen2021extract}. In addition, recent advancements \citep{Sun2019PatientKD,sun_mobilebert_2020,dynabert,romero2015fitnets} have extended its application by enabling the incorporation of information from intermediate layers into the student model's training. Moreover, studies \citep{ma2023llm,sanh_movement} demonstrate the effectiveness of combining pruning with knowledge distillation. %effectively improves performance while significantly reducing model parameters. 
However, these methods often entail extensive training time.

\begin{wraptable}[14]{r}{0.4\textwidth}
\resizebox{0.4\textwidth}{!}{
\begin{tabular}{lccc}
\hline
\multirow{ 2}{*}{Model}& \multirow{ 2}{*}{Size} & \multirow{ 2}{*}{PPL} & Inference    \\
&&&  Speedup \\
\hline
LLaMA-2 & 7B & 5.11 & 1$\times$ \\
%Phi-2 & 3B & 8.69 & 1.24× \\
%SparseGPT 2:4 & 3B & 10.17 & 1.24$\times$ \\
% Wanda 2:4 &3B & 8.34 & 0.75× \\
Wanda 2:4  &3B & 10.52 & 1.14$\times$ \\
% Bonsai & 3B & 8.89 &1.58×\\
Bonsai  & 3B & 19.47 &1.58$\times$ \\
LLM-pruner  & 3B & 19.24 &1.5$\times$ \\
\hline
% \rowcolor[gray]{0.8}
% \textbf{Faster-VTrans}  & 3B & 8.62 & \textbf{1.76x}\\
\rowcolor[gray]{0.8}
\textbf{Faster-VTrans}  & 3B &  11.8 & \textbf{1.82$\times$}\\
% \rowcolor[gray]{0.8}
% \textbf{Faster-VTrans w/o FT}  & 3B &  14.19 & 1.76x\\
\hline
\end{tabular}}
\caption{Pruned models evaluated on Wikitext-2. Our method outperforms structured pruning (Bonsai and LLM-pruner) and achieves faster inference than semi-structured pruning (Wanda 2:4). }\label{tab:llama_prune}
\end{wraptable}

\textbf{Variational Information Bottleneck.} It approximates \citep{alemi2016deep} the information bottleneck principle \citep{tishby_2000,tishby2015deep}, focusing on extracting relevant information from input variables for output variables.% aiming to reduce inter-layer information redundancy. 
It aims to maximize mutual information between intermediate layers and outputs while minimizing inter-layer mutual information to eliminate redundancy in information. 
It has been successfully applied to remove neurons in CNN, linear architectures \citep{dai2018compressing} and to RNN models \citep{srivastava2021variational}. \citet{henderson2022variational} formulate a non-parameteric variational autoencoder with VIB for transformers, but it is not aimed at pruning weights. In our work, we apply the principle for transformer compression. %Furthermore, VIB has been employed for low-resource fine-tuning in transformers \citep{mahabadi2021variational}, suggesting potential synergy with our proposed method.
% \begin{wraptable}[11]{t}{0.45\textwidth}
% \resizebox{0.45\textwidth}{!}{
% \begin{tabular}{lccc}
% \hline
% Model & Size & PPL & Inference Speedup  \\
% \hline
% LLaMA-2 (teacher) & 7B & 5.11 & 1x \\
% Phi-2 & 3B & 8.69 & 1.24× \\
% Wanda 2:4 &3B & 8.34 & 0.75× \\
% Bonsai & 3B & 8.89 &1.58× \\
% Bonsai w/o FT & 3B & 19.47 &1.58× \\
% \hline
% \rowcolor[gray]{0.8}
% \textbf{Faster-VTrans w FT}  & 3B & 8.62 & 1.76x\\
% \rowcolor[gray]{0.8}
% \textbf{Faster-VTrans w/o FT}  & 3B &  14.19 & 1.76x\\
% \hline
% \end{tabular}}
% \captionof{table}{Performance of models pruned by 50\% from LLaMA-2-7B evaluated on Wikitext-2.}\label{tab:llama}
% \end{wraptable}

%\textbf{Explainability of models.}
% Recent research has focused on understanding the learning processes of transformers with input data \citep{gulordava-etal-2018-colorless,marvin2018targeted} and model output \citep{blevins2018deep}.
% Classifiers have been employed to unveil underlying syntactic structures in transformer representations \citep{tenney2019you}. Moreover, visualizing attention heads has provided insights into the linguistic behavior of these models \citep{clark2019does,burns2018exploiting}. 
% In our comparative analysis, we focus on attention weights of pruned models, demonstrating the effectiveness of pruning in removing unnecessary information while maintaining overall performance.
\section{Background}
%\textbf{Notations.}
%$ \boldsymbol{H}$ denotes the intermediate hidden state representation. $ \boldsymbol{Y} \in \{1,2,...,n\}$ denotes the output vector. $\mathcal{D}$ denotes the joint distribution between $ \boldsymbol{X}$ and $ \boldsymbol{Y}$.
\subsection{Basic Transformer Architecture}
Transformer network $f(\cdot ; \theta_s)$ parameterised by $\theta_s$ used for NLP tasks typically consist of an embedding module and \(L\) layers with each layer containing Multi-Head Attention (MHA) module and two Feed Forward Networks (FFN). 
We denote the input to a transformer network as \(\boldsymbol{x} \in \mathbb{R}^d\) and the associated label as \(\boldsymbol{y} \in \mathcal{Y}\). Further, we represent the embedding layer hidden states obtained after the addition of positional and input embeddings as \(\boldsymbol{m}\), output of MHA activations as \( \left\{\boldsymbol{a}_i\right\}_{i=1}^L\) 
and FFN layer embedding as \(\left\{\boldsymbol{h}_i\right\}_{i=1}^L\). %The loss function is defined as $\mathcal{L}_t(\theta_s)=\mathbb{E}_{x \sim \mathcal{X}}[\ell(f(x ; \theta_s))]$ for $l$ task.
\subsection{Compressed Representations with Variational Information Bottleneck}\label{sec:VIB}
\citet{tishby2015deep} and \citet{dai2018compressing} conceptualised successive layer representations in a deep neural network to form a Markov chain while treating the input as a stochastic variable. In the context of transformers, we extend it to the successive output representations from the embedding layer, Multi-Head Attention (MHA) layers, and Feed-Forward Network (FFN) layers. Our objective is to obtain condensed intermediate representations or activations \(\tilde{\boldsymbol{k}_{i}}\) after each module (embedding layer, MHA, FFN) in the transformer architecture while preserving essential information in the predicted output $\boldsymbol{\tilde{y}}$. Unlike the compressed FFN representation as formulated in ~\cite{dai2018compressing}, $\boldsymbol{k_i} \in \mathbb{R}^{n\times seq \times d}$, $\Tilde{\boldsymbol{k_i}} \in \mathbb{R}^{n\times seq \times d^{'}}$ are each sets of vectors with $n$ input examples and $seq$ tokens and  $d^{'}$ represents the compressed  $d$ dimension.
As done by \citet{dai2018compressing}, we frame the optimization problem as: $
\label{e1}
\mathcal{L}_{i} = \beta_i I(\tilde{ \boldsymbol{ k}_{i}} , \boldsymbol{k}_{i-1})- I(\tilde{ \boldsymbol{ k}_{i}} , \boldsymbol{Y}) 
$.
Here, $I(\cdot)$ denotes mutual information between two random variables and $\beta_i$ is a hyper-parameter controlling the trade-off between compression and prediction accuracy. To simplify notation, we denote the compressed version $\tilde{ \boldsymbol{ k}} $ as $ \boldsymbol{ k} $.

%While Equation~\ref{e1} is generally intractable due to model complexity and the impracticality of the mutual information term,
To make the problem tractable, we invoke the variational upper bound as done by \citet{alemi2016deep},
\begin{equation}
\label{e2}
%\begin{split}
    \tilde{\mathcal{L}_{i}} =  \mathbb{E}_{ \boldsymbol{X},  \boldsymbol{Y}, \boldsymbol{k}_{i}, \boldsymbol{k}_{i-1} } \bigg[\beta_{i} \mathbb{D}_{ KL}\Big[p\big({ \boldsymbol{ k}_{i} } \mid  \boldsymbol{k}_{i-1}\big) \| q\big({ \boldsymbol{ k}_{i} } \big)\Big]  - \log q\big({ \boldsymbol{y} \mid { \boldsymbol{k}_{L} }}\big)\bigg] \geq {\mathcal{L}_{i}}
%\end{split}
\end{equation} 
Here, $q( \boldsymbol{ k}_{i} )$ and $q( \boldsymbol{y} \mid  \boldsymbol{k}_{L} )$ approximate $p( \boldsymbol{k}_{i})$ and $p( \boldsymbol{y} \mid  \boldsymbol{k}_{L} )$, respectively. We get the compressed representations by performing a dot product of the transformer activations with a random set of vectors $\boldsymbol{z_i} \in \mathbb{R}^{n \times seq \times d}$, where $\boldsymbol{z_i} = \boldsymbol{\mu_i} + \boldsymbol{\epsilon_i} \odot \boldsymbol{\sigma_i}$. $\boldsymbol{\mu_i} \in \mathbb{R}^{1 \times d}, \boldsymbol{\sigma_i} \in \mathbb{R}^{1 \times d}$ are learnable parameters, while $\epsilon_i \in \mathbb{R}^{n \times seq \times d}$ is sampled from non-parameterized $\mathcal{N}(\mathbf{0}, \boldsymbol{I})$, thus broadcasting $\boldsymbol{\mu_i}$ and $\boldsymbol{\sigma_i}$.
\begin{equation}
\label{e3}
\begin{array}{cc}
    \boldsymbol{k}_{i} =  \boldsymbol{z}_{i} \odot f_{i} (\boldsymbol{k}_{i-1}) & ;
    \boldsymbol{z_i} \in \mathbb{R}^{n \times seq \times d} 
\end{array}
\end{equation} 
$f(\cdot)$ denotes a standard sub-layer of the transformers like the value head projection or the up projection sub-layer.
%With the above definitions, the distributions in Equation~\ref{e2} may be defined as, $\small p\left(\boldsymbol{k}_i \mid \boldsymbol{k}_{i-1}\right)=\mathcal{N}\left(\boldsymbol{k}_i ; f_i\left(\boldsymbol{k}_{i-1}\right) \odot \boldsymbol{\mu}_i, \operatorname{diag}\left[f_i\left(\boldsymbol{k}_{i-1}\right)^2 \odot \boldsymbol{\sigma}_i^2\right]\right)$ , $\small q\left(\boldsymbol{k}_i\right)=\mathcal{N}\left(\boldsymbol{k}_i ; \mathbf{0}, \operatorname{diag}\left[\boldsymbol{\xi}_i\right]\right)$ and for $q\left(\boldsymbol{y} \mid \boldsymbol{k}_L\right)$ , the final layer in the transformer for a classification or regression task provides the essential structure. Plugging the distributions in Equation~\ref{e2} and simplifying them leads to,
Following \citet{dai2018compressing}, we derive the final VIB objective function as,
\begin{equation}
\label{e4}
\tilde{\mathcal{L}}_{\text{VIB}}=\sum_{i}^{L} \beta_i \sum_{j=1}^{r_i} \log \left(1+\frac{\mu_{i, j}^2}{\sigma_{i, j}^2}\right)
-\mathbb{E}_{{\mathbf{X}, \mathbf{Y},\mathbf{k}}}\left[\log q\left(\boldsymbol{y} \mid \boldsymbol{k}_L\right)\right]  
\end{equation}
where $r_i$ denotes the total hidden states or neurons within layer $i$. The expectation is approximated with the task classification layer.
Unlike the application of VIB on non-transformer-based networks as done by \cite{dai2018compressing}, compressing one token's representation in transformer-based models can inadvertently affect the representations of other tokens due to their inter-dependencies. However, by sampling a random vector with the shape $(samples,sequence\_length,dimensions)$, each token's representation can be uniquely adjusted by its own random vector. This approach potentially preserves more detailed context and dependencies, maintaining the integrity of the token-level information.
%Derivations can be found in the Appendix \ref{sec:deri}.

\section{Method}
%This section provides an in-depth explanation of how our methodology seamlessly integrates VIB-based pruning with knowledge distillation to compress structural components of Transformer models, ensuring accurate predictions.
Our approach involves two phases: \textbf{pruning} and \textbf{finetuning}.

During pruning, we systematically prune the student model, initialized from the finetuned teacher, using VIB-based pruning while distilling knowledge from the teacher model.
In the finetuning phase, we optimize the remaining model parameters and VIB masks to achieve the final compressed model with optimal performance. We elaborate on the primary method in section~\ref{sec:prim}, and discuss the \textit{Fast} and \textit{Faster} variants in sections~\ref{sec:fast}.

\subsection{VIB-based Pruning}\label{sec:prim}
Following Equation~\ref{e3}, we define compressed representations for each structural module in a transformer by multiplying the output of the module element-wise to a random vector.
%Similar to Equation~\ref{e3} we define 
The compressed embedding representation is thus defined as $
    \boldsymbol{m}=  \boldsymbol{z}_{m} \odot \text{Emb}(X,W)
$,
where \(\boldsymbol{z}_{m}\) is the random vector multiplied element-wise to the output of the embedding layer. $W$ denote the weight matrices of the embedding layer.
Similarly, for layer $i$ the compressed MHA representation is defined as $
\label{eqMHA}
    \boldsymbol{a}_{i}= \boldsymbol{z}_{a_{i}}\odot [\operatorname{Att}_1, \operatorname{Att}_2,..\operatorname{Att}_J]
$
%$\boldsymbol{z}_{a_{i}}$ is the random vector for MHA of layer $i$.
, compressed FFN representation as $
\label{eqFFN}
    \boldsymbol{h}_{i} =[\operatorname{gelu}\left(X W_{iU}\right) \operatorname{diag}(z_{inter,i}) \cdot W_{iD}] \odot \boldsymbol{z}_{out,i}
$,
where $W_{iU} \in \mathbb{R}^{r_i \times r_f} \text { and } W_{iD} \in \mathbb{R}^{r_f \times r}$ are intermediate up-projecting and final down-projecting layer matrices respectively. \\[1mm]
\textbf{Layer pruning.}
We incorporate the ability to prune whole MHA and FFN layers by introducing additional random vector $z_{layer}$. This enables effective pruning in scenarios with high sparsity requirements. Thus, the MHA and FFN representation for layer $i$ are defined as
$
   MHA_{{i}}=z_{layer,i}.(\boldsymbol{a}_i.W_i^{O});  \; \; \;  \;    FFN_i= z_{layer,i} \cdot \boldsymbol{h}_i
$,
 where $W_i^{O} \in \mathbb{R}^{r_i}$ denotes the output attention matrix. % while $\boldsymbol{a}_i$ and $\boldsymbol{h}_i$ are compressed representations defined previously.%in Equation~\ref{eqMHA} and Equation~\ref{eqFFN}. 

\textbf{Implementation of masks for pruning.} The $\mu_{i,j}$ and $\sigma_{i,j}$ parameters of the $z_{i,j}$ vectors of layer $i$ and neuron $j$ are initially sampled from normal distributions and iteratively refined during pruning. As shown by ~\citet{dai2018compressing}, at the minima of the Equation~\ref{e3}, whenever $\alpha_{i,j}=\mu_{i,j}^2/\sigma_{i,j}^2 = 0$, then that neuron $j$ of layer $i$ is redundant. In the inference phase, these $z_{i,j}$ vectors are converted to hard masks $z_{i,j}^{mask}$ using a thresholding operation on $\alpha_{i,j}$ for pruning neurons as defined by, 
\begin{align}
    z_{i,j}^{mask}
    & = \begin{cases}
         0 & \text{if } \log \frac{\mu_{i,j}^2}{\sigma_{i,j}^2} \leq \tau \\
         1 & \text{if } \log \frac{\mu_{i,j}^2}{\sigma_{i,j}^2} > \tau \\
    \end{cases}
\end{align}
%In our experimental setup, we set the hyperparameter $\tau$ to zero.
\textbf{Sparsity Control.}
Using $z_{i,j}^{mask}$ hard masks, we calculate model sparsity as the ratio of pruned parameters to the initial count. For FLOPs constraints, sparsity represents the reduction ratio in FLOPs to the initial value. We adopt an approach akin to ~\citet{cofi2022,dutta2023search} to incorporate a Lagrangian term which enhances stability and provides finer control over pruning by enforcing an equality constraint $s_e = t$ and introducing a violation penalty as $
    \mathcal{L}_{s}= \lambda_1 \cdot ( s_{e}- t) + \lambda_2 \cdot ( s_{e}- t)^2
$, 
where $s_e$ is the expected model sparsity, $t$ is the target sparsity and $\lambda_1, \lambda_2 \in \mathbb{R}$ are Lagrangian multipliers jointly updated during the pruning process. 
We compute $s_e$ by employing the sigmoid differentiable form of $z_{i,j}^{mask}$. 
%defined as, 
% \begin{equation}
%     %z_i^{(s)} = \text{sigmoid} \Biggl( \frac{1}{\beta} \log\Biggl( \frac{\mu_i^2}{\sigma_i^2} \Biggr) \Biggr)
%     \hat{z}_{i, j}^{\mathrm{mask}}=\sigma\left(\gamma \cdot\left(\log \frac{\mu_{i, j}^2}{\sigma_{i, j}^2}-\tau\right)\right)
% \end{equation}
% where $\gamma$ controls the sharpness of the sigmoid function. 

\begin{table*}[t]
\centering
\resizebox{\textwidth}{!}{
\begin{tabular}{l|c|cccccccc|c|c}
\hline
Model &Pars  & SST-2& QNLI &MNLI &QQP &CoLA& RTE &STS-B& MRPC & Avg & Prune Time \\%& SQuAD & Training Time \\
&(Mil) & (Acc)&(Acc)&(Acc)&(Acc)&(M. Corr)&(Acc)&(P. Corr)&(Acc) & Score & (GPU hrs)  \\%& (F1/EM) & (GPU hrs) \\
% &&&&&&&&&&&(GPU hrs)\\
\hline
BERT$_{base}$(teacher) &109 &93.1 &91.5 &84.5 &91.1 &61.2 &79.8& 89.3 &88.3 &84.8 & -\\%& 88.4/-  & -\\
DynaBERT \citep{dynabert} &  34  & 89.4 & 83.9 &76.3 & 88.8 & 15.8 & 66.4 & 85.4 & 84.6 &73.8 & $\sim$ 15\\%&-/- & $\sim$20\\
BlockPrune \citep{lagunas-etal-2021-block} & 28 & 89.3 & 84.7 &78.8 & 88.8 & 15.8 & 66.4 & 85.4 & 81.9 &73.9 &$\sim$20 \\%&-/- & $\sim$20\\
CoFi \citep{cofi2022} &28 & 90.6& 86.1& 80.6 &90.1 &35.6& 64.7 &83.1& 82.6 &76.7  &$\sim$33\\%&82.6/- & $\sim$20 \\
%PostPrune~\citeyear{kwon_fast_2022} &&&&&&&&&\\
%TinyBERT4 w/o GD~\citeyear{jiao2019tinybert}&14.5& 87.7 &81.8& 78.7 &89.5& 16.6 &47.3& 17.8 &68.9 \\%&-/- & $\leq$10 \\
 PostPrune \citep{kwon_fast_2022} &28& 75.7&63.8& 45.5& 67.5& 32.4& 58.6& 80.2&78.7 &62.8 & $\sim$0.08\\
 \rowcolor[gray]{0.8}
\textbf{VTrans (ours)} &28$^{*}$  & \textbf{92.2}&\textbf{89.7} & \textbf{83.8}& \textbf{90.5} &\textbf{43.3}& \textbf{73.4}&87.9& \textbf{87.0} & \textbf{81.0}&$\sim$20 \\%& \textbf{83.2/74.4} &$\sim$20\\
% \textbf{Ours} &14$^{*}$ & & \textbf{91.4}&\textbf{86.7} & \textbf{80.4 }& 90.0 & \textbf{35.2}& \textbf{71.8}&\textbf{83.7}& \textbf{87.5} &$\sim$20 \\%& \textbf{83.2/74.4} &$\sim$20\\
\rowcolor[gray]{0.8}
\textbf{Fast-VTrans (ours)} &28$^{*}$ &  90.8&85.8 &79.3 &89.2  &37.6 &68.0& \textbf{87.9}&83.8 & 77.8 &$\sim$ \textbf{2}  \\
\rowcolor[gray]{0.8}
\textbf{Faster-VTrans (ours)} &28$^{*}$ &  82.3&81.8 &72.1&81.4  &36.3 &67.6& 86.8&82.5 & 73.9 &$\sim$ \textbf{0.7}  \\
\hline

RoBERTa$_{base}$(teacher)  &124 & 95.3& 93.2 &87.7& \textbf{73.8}& 62.0 &78.7&91.2&90.1   &84.0&-\\
%RoBERTa$_{small6}$  &&&83.1&&&68.4&86.8&87.5 \\
MINILMv2$_6$ \citep{wang2020minilm}  &82  &  93.5 & 91.6 &84.3& 72.8& 57.8 &72.1&87.5&88.2  &81.0&$\sim$300 \\
FeatureCorr \citep{huang2023towards}  &82 &  93.8 & 92.0 &85.6&73.5 &60.3 &72.5&88.3&89.6 &82.0 &-\\
\rowcolor[gray]{0.8}
\textbf{VTrans (ours)} &82$^{*}$ & \textbf{94.8}& \textbf{93.0}&\textbf{87.3}& 73.6& \textbf{61.0}&\textbf{73.1}&\textbf{90.9}&\textbf{90.1} & \textbf{83.0} &$\sim$25 \\
\rowcolor[gray]{0.8}
\textbf{Fast-VTrans (ours)} &82$^{*}$ &  94.1&91.9 &85.4&72.9  &60.4 &72.7& 89.6&89.8 & 82.1 &$\sim$ \textbf{2.8}  \\
\rowcolor[gray]{0.8}
\textbf{Faster-VTrans (ours)} &82$^{*}$ &  93.4&90.6 & 84.8 & 72.3  & 59.2 & 72.6& 87.9&88.5 & 81.2 &$\sim$ \textbf{1}  \\
% \textbf{Ours ( \textit{Faster})} &82$^{*}$ &  \textbf{}&\textbf{} & \textbf{}&  & &&& &  &$\sim$\textbf{1} \\
\hline
GPT2$_{small}$(teacher)  &125 &92.7 &88.5&82.3 & 89.7 &47.4 & 74.4& 88.9 &85.8  &81.2&- \\
DistilGPT2 \citep{sanh2020distilbert} &82 &  90.7 & 87.9 &81.6& 66.8& 39.4&68.3&79.6&87.9 &75.3 &$\sim$ 300 \\
FeatureCorr \citep{huang2023towards} &82  & 92.0 & 88.5 &\textbf{83.4}& 70.6 & 42.3&70.2&81.6&\textbf{88.4}  &77.1&-  \\
\rowcolor[gray]{0.8}
\textbf{VTrans (ours)}  &82$^{*}$  &\textbf{92.4} & \textbf{88.8} & 82.8&\textbf{89.5} &\textbf{45.4} & \textbf{72.9}&\textbf{88.1}& 86.0   &\textbf{80.7}&$\sim$25 \\
\rowcolor[gray]{0.8}
\textbf{Fast-VTrans (ours)}  &82$^{*}$  &92.0 & 88.4&82.8& 88.7& 43.5&71.4 &86.5&86.0  &80.0&$\sim$\textbf{2.8}   \\
\rowcolor[gray]{0.8}
\textbf{Faster-VTrans (ours)}  &82$^{*}$  &91.9 & 87.3&80.1&86.2&42.1& 70.8&86.5&86.0  &78.9&$\sim$\textbf{1}   \\
\hline
\end{tabular}
}
\caption{ The performance of models on GLUE dataset. $^{*}$ indicates the average number of parameters across all datasets. Average variance in performance of our models is about $\pm0.4\%$ obtained over five random initial seeds. %FeatureCorr~\citep{huang2023towards} model performances are included from the paper itself and could not be replicated due to unavailability of code and incoherent pseudo-code.
} % \citep{cofi2022,liang_homodistil_2023} 
%Results of MobileBERT are not directly comparable as it has 24 layers and the teacher is BERT large unlike the other models. 
\label{tab:glue}
\end{table*}
\subsection{Knowledge Distillation}
We distill knowledge from the original full-sized Transformer to the student model during pruning by minimizing the cross entropy between their output probability distributions $ \boldsymbol{p}_s$ and $ \boldsymbol{p}_t$ as $
\mathcal{L}_{\text{pred}}=D_{\mathrm{KL}}\left( \boldsymbol{p}_s \|  \boldsymbol{p}_t\right) 
$.
We also adopt a layer-wise distillation approach inspired by \citet{cofi2022}. This method aligns the teacher's layer with the evolving student's, defining an intermediate distillation loss function as $\mathcal{L}_{\text {layer}}=\sum_{i \in \mathcal{T}} \operatorname{MSE}\left(W_{\text {layer}}  \boldsymbol{H}_s^{m(i)},  \boldsymbol{H}_t^i\right) $
where $W_{\text {layer }}$ is a transformation matrix initialized to identity. $\boldsymbol{H}_s^{m(i)}$ and $\boldsymbol{H}_t^i$ represent the student's and teacher's hidden layer representations, respectively, for layer distillation, with $m(\cdot)$ mapping teacher layer representations to the closest student layer as
$
m(i)=\underset{j : \text {unpruned layers}}{\arg \min } \operatorname{MSE}\left(W_{\text {layer }}  \boldsymbol{H}_s^j,  \boldsymbol{H}_t^i\right)   
$.
%\begin{equation}
%\mathcal{L}_{\text {att }}=\sum_{i \in \mathcal{T}} %\operatorname{MSE}\left(  \boldsymbol{A}_s^i, % \boldsymbol{A}_t^i\right) 
%\end{equation}
Hence, the distillation loss encompasses both the layer-wise loss and the logit-based cross entropy loss with $\eta$ controlling their relative influence:
$
\mathcal{L}_{\text {distil }}=\eta\mathcal{L}_{\text {pred}}+(1-\eta)\mathcal{L}_{\text {layer }}
$.
The comprehensive loss function to be minimised during pruning is $\mathcal{L}_{\text{total}}=  \mathcal{L}_{\text {distil }} +  \tilde{\mathcal{L}}_{\text{VIB}}+ \mathcal{L}_{s} $.
%During the finetuning phase in the primary method as shown in figure~\ref{fig:1}(a), we update the un-pruned model parameters and the Variational Information Bottleneck (VIB) masks. These masks, multiplied to their respective consecutive model layers, can subsequently be removed at the conclusion of the process. The compressed model is then derived as the dense model remaining as shown in figure~\ref{fig:1}(d) after eliminating the redundant units and the associated VIB-masks. 
The steps of the our method is reiterated in Algorithm~\ref{alg:1} in the Appendix. 
% \begin{equation}
%     \mathcal{L}_{\text{total}}=  \mathcal{L}_{\text{distil}}+  \tilde{\mathcal{L}}_{\text{VIB}}
% \end{equation}

\definecolor{Gray}{gray}{0.90}
\newcolumntype{a}{>{\columncolor{Gray}}c}
\subsection{Faster variants of our method}
\textbf{Faster compression with Data Subset.} \label{sec:fast}
%We capitalize on the inherent efficiency of fine-tuning pre-trained models to expedite the compression of the primary method. 
Leveraging the valuable features and patterns learned during the initial training phase, our \textit{Fast-VTrans} approach involves training VIB-masks and refining un-masked pre-trained model parameters with a significantly reduced data requirement as shown in figure~\ref{fig:1}(b). Intuitively, this process is expected to facilitate adaptation to lower model parameters without overfitting. To achieve this, we employ a 3\% data subset for pruning and fine-tuning, balancing efficiency and adaptability. %This \textit{fast} variant of our methodology aims to optimize the compression process, ensuring judicious use of computational resources while preserving the effectiveness of the model in the face of limited data availability.

\textbf{VIB mask training.} \label{sec:faster}
Motivated by the recent advancements in post-training pruning methodologies~\citep{kwon_fast_2022}, we propose the fastest variant, \textit{Faster-VTrans} that 
%focuses on compressing pre-trained models by fine-tuning layer masks and bias parameters where available with a subset of the whole data 
freezes all pre-trained model parameters(Figure~\ref{fig:1}(c)). During pruning, only VIB masks are trained to target irrelevant elements. Additionally, all norm and bias terms (where available) are fine-tuned to compensate for information loss. This approach, using only 3\% of the data, significantly reduces compression time while maintaining performance, making it ideal for scenarios with limited time or resources.
\begin{figure*}[t]
\centerline{\includesvg[inkscapelatex=false,width=\textwidth]{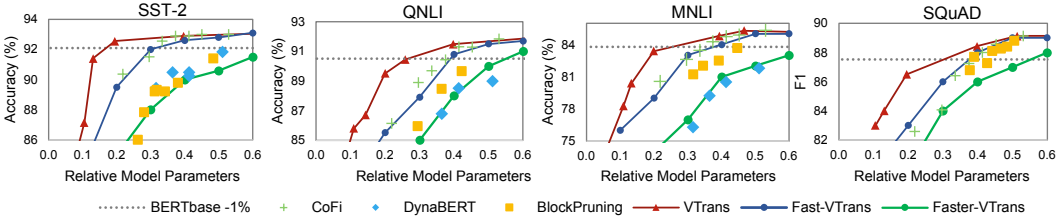}}
    %BERT_pic_1_wider.svg
    \caption{Pruning BERT-base (teacher model) on GLUE and SQuAD tasks: Each point denotes the mean performance of pruned models averaged over five random trials with different seeds with average variance of $\pm 0.4\%$ across all models. The dashed line 
    represents a 1\% performance drop from the teacher model.}
    \label{fig_bert}   
\end{figure*}

\section{Experiments}\label{sec:exp}
\subsection{Implementation Details}
\textbf{Datasets.}
%\textbf{Task-specific pruning.}
We assess the efficacy of our proposed method across a comprehensive set of linguistic understanding tasks - the General Language Understanding Evaluation (GLUE) benchmark \citep{wang-etal-2018-glue}, Stanford Question Answering Dataset (SQuAD) version 1.1 \citep{rajpurkar2016squad} and WikiText-2~\citep{merity2016pointer}. 
%\textbf{General Distillation and task-agnostic pruning.}
%For our task-agnostic experiments as shown in Table~\ref{tab:task-ag}, we utilize the open source corpus used for training BERT - Wikipedia (2500M words) and Toronto BookCorpus \citep{zhu2015aligning}, (800M words) pre-processed in a similar way as \citet{jiao2019tinybert}.Details can be found in Appendix \ref{sec:datasets}.

\textbf{Baseline Models.}
We compare our method variants with several previous techniques:  DynaBERT~\citep{dynabert}, Block Pruning~\citep{lagunas-etal-2021-block}, PostPrune~\citep{kwon_fast_2022}, CoFi~\citep{cofi2022}, FeatureCorr~\citep{huang2023towards}, DistilBERT~\citep{sanh2020distilbert}, TinyBERT-GD~\citep{jiao2019tinybert}, MiniLM~\citep{wang2020minilm}, and HomoBERT~\citep{liang_homodistil_2023}. We also compare our pruned LLaMA-2 models with previous semi-structured pruning techniques- SparseGPT~\citep{frantar2023sparsegpt}, Wanda~\citep{sun2023simple} and structured pruning techniques- LLM-pruner~\citep{ma2023llm}, Bonsai~\citep{dery2024everybody}. Details are deferred to Appendix \ref{ap:baseline}.

\textbf{Training Details.}
%\textbf{Task-specific method.}
%In our approach, we employ two phases: pruning and finetuning. For task-specific methods, the student model is initialized with the finetuned teacher on each task before pruning. We vary the sparsity constraint to achieve models with 20\% to 90\% sparsity in terms of either model size or FLOPs. 
The primary method utilizes entire datasets for pruning and finetuning, while the \textit{Fast} and \textit{Faster} variants operate on a randomly sampled subset, approximately 3\% of the data. 
%The \textit{Fast} variant trains all model parameters, while the \textit{Faster} variant focuses on VIB-mask, layer norm, and bias parameters. 
%Hyperparameters such as learning rate and the VIB-mask layers' multiplier, control the pruning-performance trade-off.
For \textit{Fast} and \textit{Faster} variants, 8000 samples are used for pruning and finetuning on large datasets and 2000 on smaller ones of GLUE and SQuAD. Finetuning uses 16000 samples for large datasets and the entire dataset for smaller ones.
%Pruning spans 20 epochs, while finetuning with the entire data requires 10 epochs; \textit{Fast} and \textit{Faster} variants use 20 epochs with smaller learning rates.
We conduct five runs with random seeds for each sparsity constraint, reporting average performance. Experiments conducted to come to the final hyper-parameter settings are provided in Appendix \ref{ap:imple_details}.
%\textbf{Task-agnostic method.}
%The pre-trained BERT model acts as the teacher in our task-agnostic method and is also initialized as the student model. We take a maximum sequence length of 128 while we prune and distil from the teacher to the student model for 3 epochs. Other hyperparameters are kept similar to ~\citep{jiao2019tinybert}. 
\subsection{Evaluation}
\textbf{Performance comparison and Compression speedup.}
In Table~\ref{tab:glue}, we compare pruned models derived from our primary method and its faster variants (\textit{Fast} and \textit{Faster}) with alternative distillation and pruning approaches. Our process starts from three teachers - BERTbase, ROBERTa, and GPT-2 small. To ensure a fair comparison of performance on GLUE datasets and compression time (measured on the largest dataset MNLI) with previous methods that share the same teacher model, we categorize the obtained models into three sections.  In the uppermost group, we compare models with 28 million parameters and 75\% sparsity to match previous smallest models~\citep{cofi2022}. Our primary method outperforms prior techniques with similar order of compression time. % similar to the state-of-the-art. 
\textit{Fast-VTrans} shows a 16x acceleration compared to CoFi, and \textit{Faster-VTrans} is at least 20x faster than DynaBERT and BlockPrune, with similar performance. Although PostPrune~\citep{kwon_fast_2022} takes lower compression time, our method yields stable performance across different seed initialization (Appendix~\ref{ap:stab}). In the middle group, our 66\% sparse model by \textit{VTrans}, outperforms both MINILMv2 and FeatureCorr, which utilize six-layer student models, with a 1\% higher GLUE score. Additionally, the faster variants demonstrate comparable performance with substantially reduced compression time. Finally, in the bottom group, when applied to the decoder-based GPT-2-small model, our compression technique demonstrates superior performance over previous approaches with minimal compression time. Furthermore, our pruned model compressed from a larger model- uncased BERT-large- outperforms DistilBERT and MINILMv2 on two GLUE tasks while achieving comparable results on the remaining tasks (Appendix~\ref{ap:bertlarge}). 

\begin{wraptable}[13]{R}{0.45\textwidth}
%\centering
\resizebox{0.45\textwidth}{!}{
\begin{tabular}{l|c|cccc}
\hline
Model &Parameters  & SST-2 &QQP & MRPC & MNLI\\%& SQuAD & Training Time \\
& (Million) &(Acc)&(Acc)&(Acc) & (Acc)\\%& (F1/EM) & (GPU hrs) \\
\hline
BERT$_{base}$ & 109& 93.1 &91.1 &88.3  &84.5 \\
TinyBERT$_4$ &14.5& 89. 7& 90.0 &  81.4 & 80.4\\% & 
MiniLM$_3$ & 17.0 & 89.3 &  88.8 &  81.9 & 78.8 \\

HomoBERT&14.1 &90.1& 89.9 &   87.3 & \textbf{81.2}\\%&\textbf{84.1/75.5} & $\sim$300\\

\rowcolor[gray]{0.8}
\textbf{VTrans w GD} &14.5 & \textbf{91.8}& \textbf{90.3} & \textbf{87.8} & 80.6\\%& 
\hline
\end{tabular}
}
\caption{The performance of models pruned using task-agnostic general distillation (GD) on GLUE: The pruned models are fine-tuned on downstream tasks. TinyBERT results are reported without data augmentation for fair comparison. \label{tab:task-ag}} 
\end{wraptable}

\textbf{Model Size and Performance comparison.}
In Figure~\ref{fig_bert}, we compare the performance of the models obtained with our task-specific approaches with previous methods across various sparsity levels in terms of relative model parameters. We note that our primary approach is able to achieve about 70 to 80\% compressed models with less than 1\% loss in accuracy from the un-pruned teacher model over all the tasks. It achieves about 20 to 70\% higher sparsity than other pruning methods like CoFi, BlockPruning and DynaBERT as it can prune embedding states along with intermediate layers, heads and hidden units unlike others. The faster variants of our method retain performance comparable to previous approaches but with 10 to 20x speedup in compression as seen in Table~\ref{tab:glue}.
%Compared to DistilBERT4 and TinyBERT6, the similar sized models obtained by our method performs better.

\textbf{Task-agnostic performance.} To evaluate our primary method on task-agnostic setting, we use general distillation during pruning and obtained a compressed model of similar size as obtained by previous approaches~\citep{liang_homodistil_2023,jiao2019tinybert}. We finetune the model on each of the tasks in GLUE and report results of four of the tasks in Table~\ref{tab:task-ag}. Our model performs better on three of the four tasks than other models obtained with previous task-agnostic approaches such as TinyBERT, MiniLM and HomoBERT.

\begin{table*}[t]
\centering
\resizebox{1.0\textwidth}{!}{
\begin{tabular}{lcacacacacacaca}
\hline
 & \multicolumn{2}{c}{ QNLI (60\%)}&\multicolumn{2}{c}{MNLI (60\%)}& \multicolumn{2}{c}{ QNLI (95\%)}&\multicolumn{2}{c}{MNLI (95\%)} &\multicolumn{2}{c}{SQuAD (60\%)}&\multicolumn{2}{c}{SQuAD  (95\%)}\\
 &Speedup & Acc &Speedup & Acc&Speedup & Acc &Speedup & Acc&Speedup & Acc&Speedup & Acc\\
\hline
Our Method &1.5$\times$ & \textbf{91.5}&1.5$\times$ & \textbf{84.3}&9$\times$ & \textbf{81.9}&9$\times$ & \textbf{74.2}&1.4$\times$ & \textbf{87.0} &7$\times$ &\textbf{ 79.6}\\
-Embeddings &1.7$\times$&90.2&1.5$\times$&83.9&9.2$\times$&80.6&9.4$\times$&73.5&1.4$\times$&84.3&7.5$\times$ &77.6\\
-Layers &1.5$\times$&91.5&1.5$\times$&84.2&7$\times$&81.9&7$\times$&73.8&1.4$\times$&85.8 &5$\times$&78.0\\
-Embeddings \& Layers&1.5$\times$&90.2&1.5$\times$&83.9&6$\times$&80.4&6$\times$&72.6&1.4$\times$ &84.5&5$\times$&75.8\\
%-Embeddings \& Heads&3x&89.7&3x&82.6&10x&79.4&10x&71.5&2x &82.9&7.5x&75.2\\
%-Embeddings,Layers, \& Heads &&&&\\
\hline
\end{tabular}
}
\caption{Investigating Pruning Strategies in BERT for varied sparsity constraints in GLUE and SQuAD Tasks: Removing embedding masks results in a performance decline. Removing layer masks reduces speedup, while removing both increases pruning of hidden states and heads, ultimately leading to diminished performance and reduced speedup in high sparsity models.\label{tab:abl}
}
\end{table*}

\subsection{Feasibility of Compression with Faster Variants}
\begin{wrapfigure}[14]{L}{0.35\textwidth}
\centerline{\includesvg[inkscapelatex=false,width=0.3\textwidth]{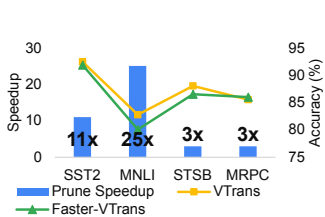}}
 \caption{Speeding-up Compression: \textit{Faster-VTrans} achieves significant speed-up over \textit{VTrans} on GLUE tasks, with minimal performance loss for models 35\% smaller than GPT-2$_{small}$.}\label{fig:speedup}
\end{wrapfigure}
% In this section, we delve into the trade-off in performance due to the usage of a reduced number of samples in the faster variants of our method. Additionally, we explore the stability in performance of the  \textit{Faster} variant, which demonstrates an impressive speed improvement of at least 20 times compared to previous approaches.\\

\textbf{Influence of data-subset on performance.}
To explore the relationship between data sample size and compression performance, we conducted pruning experiments across GLUE tasks.
%, as depicted in Figure~\ref{fig:sampTime}. 
In our trials, we used 8000 samples for larger datasets like MNLI, QQP, SST2, and QNLI, and 2000 samples for smaller datasets such as MRPC, STSB, CoLA, and RTE. Interestingly, increasing sample size for larger datasets in the faster variant led to a 1\% performance boost, but at the expense of compression times 2 to 4x longer (Appendix ~\ref{ap:sample}).
%, as shown in Figure~\ref{fig:sampTime}. 
This suggests the viability of compressing pre-trained models with limited datasets, while maintaining crucial characteristics and preserving performance to a significant degree. 

% \textbf{Stability of the fastest  \textit{Faster-VTrans} variant.} 
% We conduct a comparative assessment between the \textit{Faster-VTrans} variant and an analogous fast pruning technique based on Fisher information~\citep{kwon_fast_2022} across various GLUE tasks, as depicted in Figure~\ref{fig:fastprunecomp}. The models are generated with different FLOP constraints, resulting in models with reduced FLOPs. Notably, for extreme compression (exceeding 50\% FLOPs reduction), our method outperforms the alternative approach, yielding models with superior performance. Furthermore, our approach demonstrates models with 2 to 10\% lower variance in performance over 10 different random seed initializations, showcasing enhanced stability, as illustrated in the figure.
% \\
\textbf{Pruning Speedup in Decoder Models.}
In order to assess the compression speedup for decoder models, we employ our primary method and the \textit{Faster} variant to compress GPT-2$_{small}$. We compare the performance and speedup with both the methods across four GLUE datasets. Notably, on large datasets such as MNLI and SST-2, as depicted in Figure~\ref{fig:speedup}, there is a substantial speedup (ranging from 11 to 25x) in compression achieved with \textit{Faster-VTrans}, while incurring less than 3\% accuracy degradation compared to our primary method.

\begin{table*}
\centering
\resizebox{\textwidth}{!}{
\begin{tabular}{lc|cccccccc}
\hline
Method & WikiText2↓ & BoolQ &PIQA &HellaSwag &WinoGrande& ARC-e &ARC-c&  Average↑ & Prune\\
&PPL&Acc&Acc&Acc&Acc&Acc&Acc&Acc&Time(hrs) \\
\hline
\textbf{Unpruned model} & 5.68 & 76.5& 79.8& 76.1& 70.1& 72.8 &47.6& 70.48 & -\\
\hline
%Wanda 2:4 &  10.8 & 1.04x& 20 mins&25\\
%Wanda-sp & 110 &1.8x & 10 mins &20\\
%FLAP &14.69 & 2x & 1.8x & 25\\
%Bonsai (w Adaptation) & structured 50\% & 10.08 & &\\
% Wanda-sp &  & &  & && & & &  \\
% LLM-Pruner & & 52.32 &59.63& 35.64& 53.20 &33.50& 27.22& 33.40 &42.13 \\
% LoRAPrune &  &51.78 & 56.90 &36.76 &53.80& 33.82& 26.93& 33.10& 41.87 \\
% \rowcolor[gray]{0.8}
% \textbf{TVA-Prune} & 36 & &  & && & & &  \\
% \hline
% Wanda-sp & 132 & 50.58 & 55.01 & 29.56 &51.78& 31.27& 23.04 & 40.21& 0.16
%  \\
LLM-Pruner\dag & 16.41&  60.28 &69.31& 47.06& 53.43& 45.96& 29.18& 45.95& 1 \\
% FLAP \ddag &31.8 & 60.21& 67.52 &40.0 &57.54 &49.66 &28.49& 50.57 & 0.3\\
LoRAPrune\dag &11.60 & 61.88 & \textbf{71.53} & \textbf{47.86} &55.01& 45.13& 31.62& 52.17& $>24$ \\
Bonsai\dag & 10.92 & \textbf{67.22} & 60.85 & 43.09 & \textbf{61.64} & 54.92& 26.28 & 52.33& 40\\
\rowcolor[gray]{0.8}
\textbf{Faster-VTrans}\dag & \textbf{10.58} &63.27 & 68.56 &42.0 & 57.38& \textbf{56.97}& \textbf{26.46}& \textbf{52.44} &\textbf{2} \\
\hline
\end{tabular}}
\caption{Zero shot performance of the compressed LLaMA-7B. \dag Finetuned with LoRA \label{tab:zeroshot}. Our method takes 10 to 20 times lower time to prune than other methods with similar performance. Method with lower compression time have much lower average performance.}
\end{table*}

\subsection{Scaling-up to prune more than billion parameters}
We apply our \textit{Faster-VTrans} variant without distillation to remove redundant units from the LLaMA-2 7 billion model. Table~\ref{tab:llama_prune} displays our model's (1.6$\times$) higher inference speedup compared to Wanda~\citep{sun2023simple}, a semi-structured pruning method yielding sparse models not suitable for most hardware. Compared to LLM-Pruner~\citep{ma2023llm} and Bonsai~\citep{dery2024everybody}, our model achieves (over 1.2$\times$) higher inference speedup with significantly better performance (11.8 vs 19.24 perplexity). Pruning takes only 2 hours on a single GPU, about half the time of Bonsai. Additionally, low-rank weight adaptation \citep{hu2021low} enables us to fine-tune all modules within the same timeframe. Fine-tuning results are in Appendix~\ref{ap:wo_ft}. We also test our 50\% pruned LLaMA on zero-shot reasoning tasks shown in Table~\ref{tab:zeroshot}. Overall, our pruned model outperforms other techniques with significantly lower prune time.
%compares our approach to previous state-of-the-art pruning methods SparseGPT~\citep{frantar2023sparsegpt}, Wanda \citep{sun2023simple}, Bonsai \citep{dery2024everybody} on the WikiText-2 dataset. Results include both with and without fine-tuning after pruning, showing higher inference speedup over previous methods with comparable performance, demonstrating the effectiveness and scalability of our approach. 
%Hyper-parameters are in the Appendix~\ref{ap:wiki}.
\subsection{Ablation Study}
\textbf{Effect of pruning different structures.}
%Conduct ablation on devices
To investigate the effect of pruning embedding states unlike other previous state-of-the-art approaches, we conduct ablation studies. On removing the choice to prune the embedding states, as seen in Table~\ref{tab:abl}, the performance decreases across all tasks. This is attributed to the the increased pruning of other model structures - the layers, hidden states and the attention heads. On not pruning the layers, the models with higher sparsity levels suffer most in terms of speedup while performance degrades slightly. When both the embeddings and layers are not pruned, the algorithm prunes more attention heads and hidden states leading to lower performance of models at 60\% sparsity levels with higher sparsity (95\%) levels affected most in terms of both speedup and performance. %When embedding and attention heads are not pruned, then there is more pruning of layers and hidden states across all sparsity levels leading to lower performance and similar speedup.

\begin{wraptable}[9]{r}{0.5\textwidth}
\centering
\resizebox{0.5\textwidth}{!}{
\begin{tabular}{lcccccc}
\hline
 & \multicolumn{2}{c}{ QNLI}&\multicolumn{2}{c}{MNLI}&\multicolumn{2}{c}{SQuAD}\\
 & (60\%)  & (95\%) & (60\%) & (95\%)& (60\%) & (95\%)\\
\hline
Our Method& \textbf{91.5} &\textbf{81.9}& \textbf{84.3}& \textbf{74.2} & \textbf{87.0} &\textbf{ 79.6}\\
-$\mathcal{L}_{\text {layer }}$ &90.2&80.4&83.4&73.6&86.5&76.3\\
-$\mathcal{L}_{\text {pred }},\mathcal{L}_{\text {layer }}$ &89.8 & 79.6 &82.9 &72.4&86.2&75.5\\
%-$\mathcal{L}_{\text {atten }}$ &91.2& 80.8&84.0&72.7&86.5&77.2\\
\hline
\end{tabular}}

\caption{Ablation Experiments with and without distillation losses at different model sparsity levels.  \label{tab:ablDis}}
\end{wraptable}
\textbf{Importance of distillation.}
We analyze the impact of the distillation objectives in conjunction with our VIB-based pruning approach on pruned model performance in Table~\ref{tab:ablDis}.
We note that ablating layer-wise knowledge distillation from dynamically matched teacher layers to student layers results in a performance drop of 1 to 3 points. Completely ablating knowledge distillation leads to a performance drop of 2 to 4 points across all datasets. This shows that distillation helps improve the pruning performance of our VIB-based method by retaining performance of the student models during the iterative pruning process.
\begin{figure*}[t]
\centerline{\includesvg[inkscapelatex=false,width=0.9\textwidth]{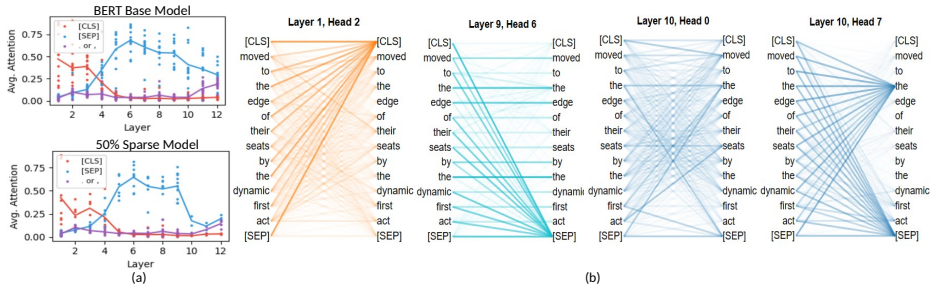}}
    \caption{(a) Comparative attention allocation towards special tokens in un-pruned (top) and pruned (bottom) models. Dots represent token attention by heads, with lines indicating mean attention by remaining heads. Pruned model exhibits reduced special token attention (b) Eliminated heads example: those highly attentive to CLS, SEP, current token, common articles like 'the' and displaying broad attention. Line thickness reflects attention weight towards the token.}
    \label{fig_plot}
    % \vspace{-3mm}
\end{figure*}

\section{Qualitative Analysis}
\textbf{Computing Average Attention in Heads.}
%what is pruned out in similar performing model?- which grammatical nuances are irrelevant?
%which are relevant?? \\
%fig- Entropy or correlation reduces across all layers \\
We evaluate information preservation and redundancy reduction in pruned models while maintaining performance by analyzing attention heads in both pre-trained BERT and an equivalently performing pruned model. On the SST-2 downstream task, we compare the un-pruned BERT base model with a 50\% smaller sized model. Our examination includes computing average attention directed towards special tokens (SEP, CLS, periods, and commas) as done by ~\citet{clark2019does}. In Figure~\ref{fig_plot}(a), we observe significant attention allocation to special tokens in the un-pruned model. ~\citet{clark2019does} speculate that attention to these tokens might serve as 'no-ops' when a head's function is not applicable, making such heads viable candidates for pruning. In the 50\% pruned model, we confirm this observation, noticing a reduction in attention heads in the latter layers (10-12) and a drastic decrease in average attention towards these tokens in the remaining heads. Additionally, we observe a shift in average attention in each layer from the current token to the previous and next token in the pruned model (see Appendix~\ref{ap:att_tok}).\\
\textbf{Redundancy in Attention Heads.}
We visualize attention heads\footnote{\url{https://github.com/jessevig/bertviz}} to identify redundant and preserved language aspects in pruned models. %For this, we display attention weights for selected SST-2 dataset examples. 
In Figure~\ref{fig_plot}(b) displaying attention weights, we observe that heads with high focus on CLS or SEP tokens and excessive attention to the current token are often pruned. %, based on the hypothesis that such attention may not provide novel information. 
Similarly, heads with broad attention or a focus on common articles like 'the' or 'a' are removed. %Retained heads in the pruned model involve those which attend to next or previous tokens or focus on key words determining task performance. 
Moreover, we observe that our method aligns with the interpretation by ~\citet{burgess2018understanding} suggesting that VIB encourages the acquisition of more disentangled representations (see Appendix~\ref{ap:dis}).

\section{Conclusion}
Integrating the Variational Information Bottleneck principle, our structured pruning method efficiently prunes redundant units across all structural levels in transformers. This results in upto 70\% higher compression compared to prior methods, accompanied by minimal accuracy loss ($<$1\%) on GLUE and SQuAD tasks, along with up to 4x inference speedup. Unlike previous approaches, we address both parameter and FLOPs constraints with our task-specific distillation approach, requiring significantly lower training resources (upto 60x less) than task-agnostic methods while achieving similar performance. Our efficient compression variants, namely \textit{Fast} and \textit{Faster-VTrans}, deliver over 10x pruning-speedup with performance comparable to previous state-of-the-art approaches. Our evaluation encompasses various architectures, including decoder-based GPT models. Demonstrating the effectiveness and scalability of our method, we compress the large LLaMA-2 model by 50\% with better performance (7 points) and enhanced inference speed (1.2$\times$) than other structured pruning methods with lower compression time. Our qualitative analysis supports the competitiveness of our approach, with potential for further quantification in future work.


% \section{Ethics Statement}
% Authors can add an optional ethics statement to the paper. 
% For papers that touch on ethical issues, this section will be evaluated as part of the review process. The ethics statement should come at the end of the paper. It does not count toward the page limit, but should not be more than 1 page. 

\bibliography{colm2024_conference}
\bibliographystyle{colm2024_conference}
\newpage
\appendix

\begin{algorithm}[t]
\caption{ Our primary method variant - Identifying redundant units with VIB-based masks with user-defined parameters or FLOPs constraint}\label{alg:1}
\begin{algorithmic}
\State \textbf{Input:} Pruning metric (parameters or FLOPs), target metric $t$, Teacher Model
%\State \textbf{Initialize:}Student Model $f(\cdot ; \theta_s)$
\State \textbf{Initialize:}VIB masks $z_m,z_{a},z_{out}, z_{layer}$ %-VIB masks
%\State $\hat{z}_{i, j}^{\mathrm{mask}}=\sigma(\gamma \cdot[\log \frac{\mu_{i, j}^2}{\sigma_{i, j}^2}-\tau])$ \Comment{\#differentiable masks}
\For{$e={1,...,Epochs }$}
\State Calculate VIB loss $\tilde{\mathcal{L}}_{\text{VIB}}$
%= \sum_{i}^{L} \beta_i \sum_{j=1}^{r_i} \log (1+\frac{\mu_{i, j}^2}{\sigma_{i, j}^2}) $ \Comment{\#VIB loss}
\State Calculate expected sparsity $s_e$% = \sum_{i \in L,j \in d_L} \sum_{masks}\hat{z}_{i, j}^{\mathrm{mask}} $ %\Comment{\#calculate expected sparsity}
\State Calculate sparsity loss $ \mathcal{L}_{s} $ %= \lambda_1 \cdot ( s_{e}- t) + \lambda_2 \cdot ( s_{e}- t)^2$ \Comment{\#sparsity loss}
%\State Calculate distillation loss $\mathcal{L}_{\text {distil }}$ %=\eta\mathcal{L}_{\text {pred}}+(1-\eta)\mathcal{L}_{\text {layer }}$ \Comment{\#distillation loss- combining layer and prediction losses }
\State Total loss $\mathcal{L}_{\text{total}}=   \mathcal{L}_{\text{distil}} +
 \tilde{\mathcal{L}}_{\text{VIB}}+ \mathcal{L}_{s} $ %\Comment{\#total loss}
%\State $M_{VIB} \gets M \odot \hat{z}_{i, j}^{\mathrm{mask}}$ \Comment{\#update model weights}
\State Update model $f(\cdot ; \theta_s)$ weights $\theta_s$
\State Update $z$ masks 
\State Update $\lambda_1, \lambda_2$
\EndFor
%\State \#To finetune model $M_{VIB}$ with masked units
\State Convert $z$ masks to binary masks $\hat{z}_{i,j}^{mask}$%$= 0$ if $\log \frac{\mu_{i,j}^2}{\sigma_{i,j}^2} \leq \tau$ else $1$
%\Comment{\#prune VIB mask-selected units}
\State Fine-tune VIB mask-selected model weights by minimizing $\mathcal{L}_{\text{total}}=  \mathcal{L}_{\text{distil}}+\tilde{\mathcal{L}}_{\text{VIB}}$
\State Prune VIB mask-selected units from model
\State $\hat{f}(\cdot ; \theta_s) \gets f(\cdot ; \theta_s)$ $\odot \hat{z}_{i,j}^{mask} $ 
%\For{$e={1,...,Epochs}$}
%\State $\mathcal{L}_{\text{total}}=  \mathcal{L}_t +\mathcal{L}_{\text{distil}}$ %\Comment{\#total loss- with target loss and distillation loss}
%\State Update model $\hat{f}(\cdot ; \theta_s)$ weights 
%\EndFor
\State \textbf{Output:} Compressed model $\hat{f}(\cdot ; \theta_s)$
\end{algorithmic}
\end{algorithm}

\section{Further Training details}\label{sec:trn}
\subsection{Datasets}\label{ap:datasets}
\textbf{General Language Understanding Evaluation (GLUE)} \citep{wang-etal-2018-glue} is a collection of nine natural language understanding tasks. The GLUE tasks encompass various domains, including sentiment analysis (SST2, \citep{socher-etal-2013-sst2}), natural language inference (MNLI, \citep{mnli}), paraphrase identification (QQP
%\footnote{\url{https://quoradata.quora.com/First-Quora-Dataset-Release-Question-Pairs}} 
and QNLI), textual entailment (MRPC, \citep{dolan-brockett-2005-mrpc}), linguistic acceptability (CoLA \citep{warstadt-etal-2019-cola}), semantic textual similarity (STS-B \citep{cer-etal-2017-stsb}), and recognizing textual entailment (RTE \citep{rte1,rte2,rte3,rte5}).

\textbf{Stanford Question Answering Dataset (SQuAD)} \citep{rajpurkar2016squad} is a reading comprehension dataset, consisting of questions posed on Wikipedia articles, where the answer are a segment of the comprehension text.

\textbf{WikiText-2}
Created from Wikipedia articles~\cite{merity2016pointer}, this dataset consists of 2 million tokens for training and more than 200,000 tokens in the validation and test set.

\begin{table*}[t]
\centering
\resizebox{1.0\textwidth}{!}{
\begin{tabular}{l|cccc}
\hline
Method & Pruning Setting & Pruning Granularity & Distillation Setting & Student Initialization \\
\hline
DistilBERT~\citep{sanh2020distilbert}&no pruning&-&TA&Un-trained weights\\
TinyBERT-GD~\citep{jiao2019tinybert} &no pruning&-&TA&Un-trained weights\\
MiniLM~\citep{wang2020minilm}&no pruning&-&TA&Un-trained weights\\
HomoDistil~\citep{liang_homodistil_2023}&prune while distill&Row, Column&TA&pre-trained weights\\
DynaBERT~\citep{dynabert}&prune, then distill&FFN, Head&TS& Pruned, Pre-trained weights\\
Block Pruning~\citep{lagunas-etal-2021-block}&prune, then distil&weight blocks &TS&Pruned, Pre-trained weights\\
Post-prune \citep{kwon_fast_2022}&prune&MHA, FFN&none &-\\
CoFi\citep{cofi2022} &prune while distill&Head, FFN, Layers&TS&Pre-trained weights\\
\hline
\rowcolor[gray]{0.8}
Our method &prune while distill& Embeddings, Head,&TA/TS&Pre-trained/finetuned\\
\rowcolor[gray]{0.8}
&&FFN, Layers &&teacher weights\\
\hline
SparseGPT~\citep{frantar2023sparsegpt}&prune&MHA, FFN&none &-\\
Wanda~\citep{sun2023simple}&prune&MHA, FFN&none &-\\
LLM-Pruner~\citep{ma2023llm} &prune&MHA, FFN&none &-\\
Bonsai~\citep{dery2024everybody}&prune&MHA, FFN&TS &Pre-trained compressed model\\
\hline
\rowcolor[gray]{0.8}
Our method (for LLM pruning) &prune & Embeddings, Head,&TS&Pre-trained \\
\rowcolor[gray]{0.8}
&&FFN, Layers &&compressed model\\
\hline
\end{tabular}}
\caption{Comparison of the baseline methods with our method in terms of pruning and distillation techniques. TS - indicates task specific; TA - Task agnostic distillation \label{tab:base}}
\end{table*}
\subsection{Further details about the baseline models}\label{ap:baseline}
In Table~\ref{tab:base}, we present a comprehensive comparative analysis of the various baseline methods used for performance evaluation in Section \ref{sec:exp}.

\textbf{DistilBERT} \citep{sanh2020distilbert} employs a vanilla distillation approach during the pre-training phase by considering the distillation loss over the outputs of the models. \textbf{TinyBERT} \citep{jiao2019tinybert} builds upon DistilBERT by leveraging knowledge from intermediate Transformer layers. \textbf{MiniLM} \citep{wang2020minilm} targets the discrepancy between queries-keys scaled dot product and values-values scaled dot product in the final layer's self-attention module. Extending on MiniLM, MiniLMv2 \citep{wang2020minilmv2} emphasizes on the emulation of attention head relations from the teacher model. Employing a task-agnostic distillation approach, \textbf{HomoDistil} \citep{liang_homodistil_2023} distils knowledge into iteratively pruned students initialized to the teacher's weights. \textbf{DynaBERT} \citep{dynabert} offers flexibility to adjust the model size and latency by selecting an adaptive width and depth. The training involves an initial training of a width-adaptive model followed by enabling adaptability in both width and depth, using distillation from the full-size model. \textbf{CoFi} \citep{cofi2022} adopts an iterative pruning strategy during the finetuning phase, utilizing distillation from the larger task-specific teacher models to the student. It performs simultaneously pruning of attention heads, FFNs and whole layers. Movement Pruning \citep{sanh_movement} proposes a deterministic first-order weight pruning method, that is effective in finetuning regime of pre-trained models. \textbf{BlockPruning} \citet{lagunas-etal-2021-block} uses this method to prune blocks of weight matrices, allowing for better optimizations on dense hardware. \textbf{Post-Prune} uses post-training pruning framework \citep{aguilar2020knowledge} that proposes a Fisher-based technique to train masks for identification of redundant neurons and selectively fine-tunes only the masks to specific tasks. %Although the method has significantly lower pruning and finetuning times over other methods, the method's performance is compromised.

We show the scaling ability of our pruning technique to prune large language models like LLaMA-2~\citep{touvron2023llama2}. Comparison with previous techniques includes \textbf{SparseGPT}~\citep{frantar2023sparsegpt} which proposes a second-order layer-wise pruning method that approximates closed form equations thus being able to scale up pruning LLMs. \textbf{Wanda} ~\citep{sun2023simple} takes into account the norm of weights and input activations for pruning weights in an unstructured/structured manner. \textbf{Bonsai} ~\citep{dery2024everybody} is a gradient-free structured pruning method that estimates module importance perturbatively by generating sub-models and evaluating their performances. \textbf{LLM-pruner}~\citep{ma2023llm} is a structured pruning method that uses gradient information to prune large language models in a task-agnostic manner.
%In contrast, our method proposes a VIB-guided pruning strategy that is able to compress the embedding states along with the other structural parts of transformers. Our method is adaptable to both task-agnostic and task-specific settings.
\subsection{Influence of Dataset size on pruning and finetuning time}\label{ap:sample}
\begin{figure}[h]
\centerline{\includesvg[inkscapelatex=false,width=\textwidth]{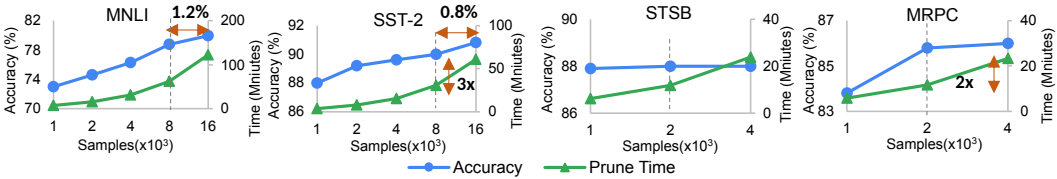}}
 \caption{Data subset impact: Prune time rises exponentially with data samples, while model performance stabilizes.}
 \label{fig:sampTime}
 \end{figure}

We prune with different number of samples from the datasets within GLUE with our \textit{Fast-VTrans} method as seen in Figure~\ref{fig:sampTime}. For both the faster variants, we use 8000 samples from large datasets (MNLI, QNLI, QQP, SST-2, SQuAD) and 2000 samples from the small datasets (MRPC, RTE, STSB, CoLA) during pruning. Further increasing the samples leads to a 2x to 4x increase in prune time with only about 1\% boost in performance.  During finetuning with the faster variants, we use 16000 samples for large datasets and the whole small datasets.

\subsection{Influence of number of epochs}\label{ap:epoch}
\begin{figure*}[h]
\centerline{\includesvg[inkscapelatex=false,width=\textwidth]{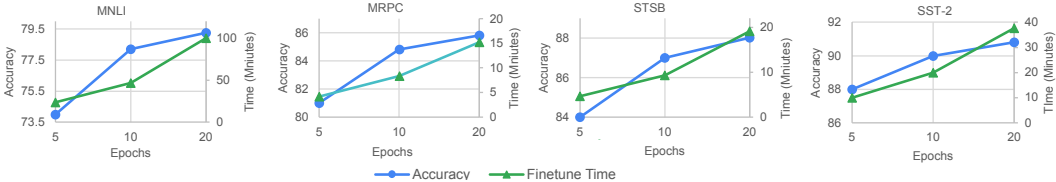}}
    \caption{We finetune the models obtained with sparsity of 75\% with different number of epochs and evaluate the final performance.} %Datasets with lower number of samples gain from larger number of epochs.}
    \label{fig_epvstime}
    % \vspace{-6mm}
\end{figure*}

The number of epochs used to finetune the model is varied from 5 to 20 as seen in Figure~\ref{fig_epvstime}. For smaller GLUE tasks, we finetune the weights of the student model with distillation for 20 epochs. For larger GLUE tasks, we finetune for 10 epochs with distillation. Increasing the number of epochs for larger datasets (MNLI, SST-2) is seen to increase the time by 2x with less than 1\% increase in accuracy.

\subsection{Implementation Details of experiments with GLUE and SQuAD}\label{ap:imple_details}
 %The code for our method is attached as a supplementary folder. 
 We list the hyper-parameter setting for task-specific pruning and finetuning with our method on GLUE and SQuAD tasks in Table~\ref{tab:hyp}. Experiments are performed on a single NVIDIA V100 GPU (32GB). In task-specific setting, we finetune the pre-trained BERT-base to particular tasks for 3 (large datasets) and 10 (small datasets) epochs to obtain teachers as used in previously~\cite{dynabert}. BERT model previously finetuned on MNLI task is utilized to get teachers of smaller GLUE datasets - RTE, MRPC and STS-B. These teacher yield high performing models.
 
For \textit{VTrans} and \textit{Fast-VTrans}: We prune and finetune with our objective function for 20 epochs to obtain smaller student models within the given sparsity constraint. Epochs lower than 20 does not let the pruning algorithm converge for higher sparsity levels(>80\%). For lower sparsity levels, the algorithm converges within less than 10 epochs. For \textit{Faster-VTrans}, we freeze the model parameters while training only the VIB masks. %We prune for 20 epochs and finetune the masks for 10 epochs. 
 
 For our task-agnostic variant, we use similar hyperparameters used by ~\citet{jiao2019tinybert} and VIB parameters are kept as in Table~\ref{tab:hyp}. We use 8 NVIDIA V100 to prune the BERT-base pre-trained model for 3 epochs. The pruned model is finetuned for 20 epochs on small and 10 the large tasks of GLUE and SQuAD as seen optimal in Figure~\ref{fig_epvstime}.
\begin{table}
\centering
\resizebox{0.7\columnwidth}{!}{
\begin{tabular}{lccc}
\hline
 Hyper-parameters  &  GLUE(small) & GLUE(large) &SQuAD  \\
 \hline
 VIB learning rate & \{1 to 2\} $\times 10^{-2}$ & \{1 to 2\} $\times 10^{-2}$ &\{1 to 2\} $\times 10^{-3}$\\
 %VIB $\beta_{layer}$ & 100 & 100 &100\\
%VIB $\beta_{embedding}$ & 25 to 50 & 25 to 50 &25 to 50\\
%Lagrangian warmpup epochs & 6 & 2 & 2 \\
 Training Learning Rate  & \{2\} $\times 10^{-5}$ &\{2\} $\times 10^{-5}$ &  \{3\} $\times 10^{-5}$ \\
 Fine-tuning Learning Rate & \{2 to 7\} $\times 10^{-5}$ &\{2 to 7\} $\times 10^{-5}$ & \{2 to 7\} $\times 10^{-5}$ \\
Batch Size  & 32 &32 & 16\\
Fine-tune Epochs & 20 &10 & 3\\
Distillation $\eta$ & 0.5 & 0.5 &0.5 \\ 
%Learning Rate Decay & Linear\\
%Learning Rate Warmup  & 0.05 \\
%Max Sequence Length  & 128\\
%Dropout of Task Layer  & 0.1\\
% Weight Decay & 0 \\
%Adam  $\beta_1$ & 0.9\\
%Adam  $\beta_2$&0.99 \\
% Adam  $\epsilon$ &1$\times 10^{6}$ \\
%Gradient Clipping & 1\\
\hline
\end{tabular}}
\caption{Hyper-parameter configurations for pruning and finetuning with our method and its variants on GLUE and SQuAD datasets. Experiments are conducted on NVIDIA V100 GPU. \label{tab:hyp}}
\end{table}

\begin{table}
\centering
\begin{subtable}[h]{0.4\textwidth}
\centering
\resizebox{0.7\textwidth}{!}{
\begin{tabular}{lcc}
\hline
VIB LR & Dataset size & block\_size\\
%  \hline
% \rowcolor[gray]{0.8}
% \multicolumn{3}{c}{Pruning} \\
\hline
5x$10^{-2}$ & 4000 & 512\\
\hline
\end{tabular}}
\caption{ \label{tab:hyp_llama_prune}}
\end{subtable}
\hfill
\begin{subtable}[h]{0.55\textwidth}
\resizebox{\textwidth}{!}{
\begin{tabular}{lcccc}
\hline
weights LR & LoRA-rank & LoRA-$\alpha$ & $\eta$ (distill Weight) & block\_size\\
%  \hline
% \rowcolor[gray]{0.8}
% \multicolumn{5}{c}{Finetuning} \\
\hline
1x$10^{-4}$ & 128 & 4$\times$rank & 0.01 & 512 \\
\hline
\end{tabular}}
\caption{ \label{tab:hyp_llama}}
\end{subtable}
\caption{Hyper-parameters for (a) pruning with Faster-VTrans and (b) for fine-tuning compressed model on WikiText-2 Dataset}
\end{table}
\begin{figure*}[t]
\centerline{\includesvg[inkscapelatex=false,width=0.9\textwidth]{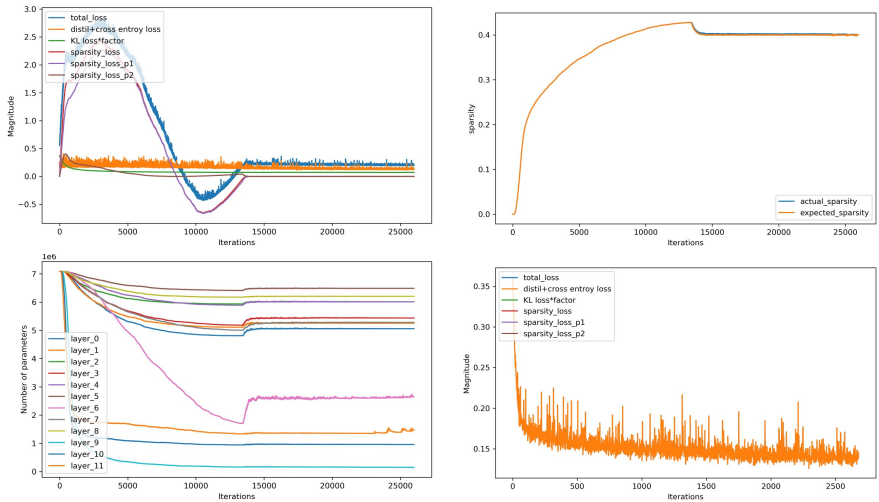}}
    \caption{Training plots for the CoLA dataset under a 40\% model size sparsity constraint. Top left: illustrates the evolution of various loss objectives over training iterations. Top right: the pruning trajectory stabilizes with increasing iterations. Bottom left: depicts the trajectory of remaining parameters in each layer. Bottom right: showcases different loss trajectories during finetuning. \label{fig:train}}
\end{figure*}
\subsection{Implementation details with experiments on WikiText-2}\label{ap:wiki}
Our pruning method \textit{Faster-VTrans} essentially only introduces a single hyper-parameter, the VIB learning rate, which we set at 5x$10^{-2}$. Increasing the VIB learning rate (LR) reduces pruning time but may result in lower model performance. Conversely, decreasing the VIB LR improves model performance but requires more data samples to converge to the user-defined sparsity level, thereby increasing pruning time. We prune with only 4000 data samples from the train set to keep our prune time within 2 GPU hours. During finetuning, we use LoRA to finetune the all modules except embeddings. The hyper-parameters are listed in Table~\ref{tab:hyp_llama}. All experiments with LLaMA are run on NVIDIA A6000 GPU (48GB).
%Before fine-tuning, the model produced by Wanda 2:4 achieves a speedup over the parent model (1.14×) but is slower and less accurate than Bonsai’s pruned model (8.89ppl vs 10.52ppl). While the performance gap can be bridged by LoRA finetuning $(10.52 \to 8.34)$, the adapted semi-structured model experiences a drastic slowdown (0.75×), since the learned low-rank matrices cannot be merged with the original sparsified ones without reverting back to dense computation. Thus, LoRA fine-tuned Wanda 2:4 is twice slower ($\sim$ 0.48×) than the model from Bonsai.Bonsai takes 40 hours to get prune and finetune
\subsection{Resources required for pruning LLMs}
The details about the number of data samples and maximum sequence length are provided in Table~\ref{tab:mem_data} below. Although Bonsai is a forward-pass-only method, it takes approximately 37 hours to prune 50\% of the model parameters. This duration is significantly longer than other forward-pass methods and our method due to Bonsai's exploration of sub-models and repeated evaluations. Bonsai processes one sample with a sequence length of 4096 in a single pass, whereas our method handles one sample with a maximum sequence length of 512 using a single NVIDIA A6000 (48GB) GPU. In contrast, LLM-Pruner requires two GPUs of 80GB each or four 40GB GPUs, which is four times the GPU memory required by our method.
\begin{table}[h]
\centering
\resizebox{\columnwidth}{!}{
\begin{tabular}{l|cccccc}
\hline
Method & Wikitext2 PPL(w/o) & Max Memory & Pruning & Sequence & Pruning  & Inference  \\
& finetune & during pruning &  Samples  & length &Time &Speedup \\
\hline
LLM-pruner &19.24 & 154 GB & 10&64 & 1 hr & 1.5x \\
Bonsai & 19.47 & 25 GB & 32 & 4096 & $~37$ hrs & 1.58x\\
\hline
Ours & 11.8 & 36 GB & 5000 & 512&  $~2$ hrs & 1.82x\\
\hline
\end{tabular}
}
\caption{Comparison of details of pruning LLaMA-2-7B model using structured pruning methods } 
\label{tab:mem_data}
\end{table}
\subsection{Pruning/Training plots}
We show the trajectory of different losses and layer parameters in our final objective function during pruning and finetuning for a particular task in Figure~\ref{fig:train}. Our sparsity constraint is seen to stabilize pruning after certain epochs.
%\subsection{Importance of controlling VIB Lagrangian parameter $\beta$}

\subsection{FLOPs and Speedup}
With FLOPs constraint in our objective function instead of model parameters, we calculate the sparsity as the reduction in FLOPs of the model to the original FLOPs. We evaluate our model using the Pytorch Profiler (\url{https://pytorch.org/tutorials/recipes/recipes/profiler_recipe.html}) and compare the inference speedup and the FLOPs (excluding the embeddings for consistency with other methods) of the baseline models with a model obtained using our method in Table~\ref{tab:flops}. Although certain methods yield models that have faster inference speed, our model can be compressed more in terms of number of parameters than the others while retaining performance. 
\begin{table*}
\centering
\resizebox{0.8\textwidth}{!}{
\begin{tabular}{c|c|c|c}
\hline
 Model & \text { Params (million) } & \text { Inference Speedup } &  \# FLOPs (non-embedding) \\
 \hline
 BERT-base  & 109 & 1.00 $\times$ & 1.00 $\times$ \\
DistilBERT$_6$ & 66 & 1.98 $\times$ & 0.50 $\times$ \\
TinyBERT$_6$- GD  & 66 & 1.98 $\times$ & 0.50 $\times$ \\
MiniLMv1$_6$ & 66 & 1.98 $\times$ & 0.50 $\times$ \\
%MiniLMv2_6 & 66 & 1.98 $\times$ & 0.50 $\times$ \\
HomoBERT-base  & 65 & 1.30 $\times$ & 0.56 $\times$ \\
BERT-small  & 28.6 & 4.77 $\times$ & 0.15 $\times$ \\
$\mathrm{CoFi}_{5 \%}$ & 28.4 & 5.16 $\times$ & 0.05 $\times$ \\
TinyBERT$_{3 \times 384}$-$\mathrm{GD}$ & 17.0 & 7.34 $\times$ & 0.06 $\times$ \\
MiniLMv1$_3$ & 17.0 & 7.34 $\times$ & 0.06 $\times$ \\
TinyBERT$_{4 \times 312}$-GD & 14.5 & 6.28 $\times$ & 0.07 $\times$ \\
HomoBERT-xsmall  & 15.6 & 2.51 $\times$ & 0.10 $\times$ \\
 HomoBERT-tiny  & 14.1 & 2.55 $\times$ & 0.09 $\times$ \\
 \hline
 Ours & 14.5  & 4 $\times$ & 0.07 $\times$ \\
\hline
\end{tabular}%
}
\caption{ We analyze and compare the inference speedup and FLOPs count (excluding embedding) for the baselines and our method, relative to BERT-base. \label{tab:flops}}
\end{table*}

\section{Further comparison with other methods}
\subsection{Comparison on WikiText-2 after fine-tuning pruned LLMs}\label{ap:wo_ft}
The performance of models after finetuning on WikiText-2 is seen in Table~\ref{tab:llama_time}. The performance of our model is slightly worse since we use the WikiText-2 dataset with \textbf{5000} data samples for fine-tuning. Whereas, Bonsai, Wanda use c4~\citep{raffel2020exploring} dataset with over \textbf{15000} samples for finetuning, thus requiring 3 times more fine-tuning time. Before fine-tuning, the model produced by Wanda 2:4 achieves a speedup over the parent model (1.14×). While the performance gap can be bridged by LoRA finetuning $(10.52 \to 8.34)$, the adapted semi-structured model experiences a drastic slowdown (0.75×), since the learned low-rank matrices cannot be merged with the original sparsified ones without reverting back to dense computation. 
SparseGPT already uses weight updates in its algorithm. 
\begin{table}[h] %[11]{r}{0.45\textwidth}
\centering
% \begin{subtable}[h]{0.45\textwidth}
% \resizebox{0.9\textwidth}{!}{
% \begin{tabular}{lccc}
% \hline
% \multirow{ 2}{*}{Model}& \multirow{ 2}{*}{Size} & \multirow{ 2}{*}{PPL} & Inference    \\
% &&&  Speedup  \\
% \hline
% % LLaMA-2 & 7B & 5.11 & 1x  \\
% SparseGPT 2:4 & 3B & 10.17 & 1.24x \\
% % Wanda 2:4 &3B & 8.34 & 0.75× \\
% Wanda 2:4 w/o FT &3B & 10.52 & 1.14× \\
% % Bonsai & 3B & 8.89 &1.58×\\
% Bonsai w/o FT & 3B & 19.47 &1.58× \\
% LLM-pruner w/o FT & 3B & 19.24 &1.6x \\
% \hline
% % \rowcolor[gray]{0.8}
% % \textbf{Faster-VTrans}  & 3B & 8.62 & \textbf{1.76x}\\
% \rowcolor[gray]{0.8}
% \textbf{Faster-VTrans w/o FT}  & 3B &  13.4 & \textbf{1.76x}\\
% \hline
% \end{tabular}}
% \caption{}\label{tab:wo_ft}
% \end{subtable}
% \hfill
%\begin{subtable}[h]{0.7\textwidth}
\centering
\resizebox{0.7\textwidth}{!}{
\begin{tabular}{lcccccc}
\hline
\multirow{ 2}{*}{Model}&Prune & \multirow{ 2}{*}{Size} & Data &\multirow{ 2}{*}{PPL} &Inference & Prune Time \\ %& Fine-tune time \\
&Method&&samples & &Speedup&  (GPU hrs) \\ %& (GPU hrs) \\
\hline
 % LLaMA-2 & 7B & 5.11 & 1x  \\
SparseGPT 2:4 &semi-structured& 3B & - & 10.17 &1.24x & 0.5 \\ %&-\\
% Wanda 2:4 &3B & 8.34 & 0.75× \\
Wanda 2:4 w FT &semi-structured&3B & 15000 &8.34 & 0.75x & 0.06 \\% & 1.5\\
% Bonsai & 3B & 8.89 &1.58×\\
Bonsai w FT &structured& 3B &15000 & 8.89 & 1.58& 35 \\%& 1.4\\
\hline
% \rowcolor[gray]{0.8}
% \textbf{Faster-VTrans}  & 3B & 8.62 & \textbf{1.76x}\\
\rowcolor[gray]{0.8}
\textbf{Faster-VTrans w FT}  & structured&3B &5000 &  9.47 & \textbf{1.82x} & 2 \\%& 1.4\\
\hline
\end{tabular}}
%\end{subtable}
\caption{Performance of models pruned from LLaMA-2-7B  evaluated on Wikitext-2 after (w) finetuning (FT). }\label{tab:llama_time}
\end{table}

\subsection{Stability of the fastest  \textit{Faster-VTrans} variant.}\label{ap:stab} 
\begin{figure}[h]
\centerline{\includesvg[inkscapelatex=false,width=0.6\textwidth]{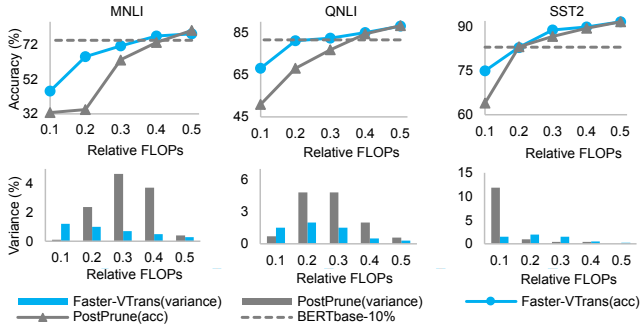}}
    \caption{Comparison: \textit{Faster-VTrans} vs. PostPrune~\citep{kwon_fast_2022} on 3 GLUE tasks. \textit{Faster-VTrans} achieves superior accuracy and stability on compression from BERT-base }
    \label{fig:fastprunecomp}   
\end{figure}

 We conduct a comparative assessment between the \textit{Faster-VTrans} variant and an analogous fast pruning technique based on Fisher information~\citep{kwon_fast_2022} across various GLUE tasks, as depicted in Figure~\ref{fig:fastprunecomp}. The models are generated with different FLOP constraints, resulting in models with reduced FLOPs. Notably, for extreme compression (exceeding 50\% FLOPs reduction), our method outperforms the alternative approach, yielding models with superior performance. Furthermore, our approach demonstrates models with 2 to 10\% lower variance in performance over 10 different random seed initialization, showcasing enhanced stability, as illustrated in the figure.
 
\subsection{Pruning BERT-large}\label{ap:bertlarge}
We also evaluated our primary approach for compressing large models such as uncased BERT-large. Our models outperform DistilBERT and MINILMv2 on two GLUE tasks while achieving comparable results on the remaining tasks as seen in Table~\ref{tab:glue_large}.
\begin{table}[h]
\centering
\resizebox{0.5\textwidth}{!}{
\begin{tabular}{l|c|cccc}
\hline
Model &Parameters  & SST-2 & MNLI & RTE & MRPC \\
& (Million) & (Acc) & (Acc) & (Acc) & (Acc) \\
\hline
BERT$_{large}$ &334 & 94.9  & 86.6  & 83.8 &  89.3  \\
DistilBERT$_6$ &66 &92.5&82.4&58.4&86.9  \\
MINILMv2$_6$ &66 &92.4&\textbf{83.4}&71.3&88.6 \\
\rowcolor[gray]{0.8}
\textbf{VTrans} &66$^{*}$ & \textbf{92.5} &83.1&\textbf{73.6}&\textbf{89.1} \\
% \textbf{Ours -  \textit{fast}} &66$^{*}$ &  & &  &  \\
\hline
\end{tabular}
}
\caption{ Compression of BERT-large model: Comparing performance of compressed models on GLUE dataset. $^{*}$ indicates the average number of parameters across all datasets. } 
\label{tab:glue_large}
%\end{table}
\end{table}
\subsection{Key differences from other iterative pruning methods}
We provide a detailed discussion of the key differences of our method from other pruning methods like HomoDistil~\citep{liang_homodistil_2023} and CoFi~\citep{cofi2022}:
 
\textbf{Flexibility in pruning and distillation settings:} In Table~\ref{tab:base} presents a comprehensive comparison of our method with other baseline techniques concerning pruning and distillation. \\
Notably, our training method exhibits flexibility in working within both task-agnostic and task-specific settings. While HOMODISTIL operates primarily in a specific setting, our approach accommodates both. Moreover, our task-specific setting demonstrates remarkable efficiency with significantly lower resource requirements, utilizing just 1 GPU for tasks such as GLUE and SQuAD, as opposed to the substantial resource demand of at least 8 GPUs and extensive training time in the broader setting.

\textbf{Performance and Resource Efficiency:} Table~\ref{tab:task-ag} showcases that our task-specific variant achieves either better or comparable performance in certain tasks compared to HOMODISTIL, despite utilizing fewer GPUs. This efficiency is crucial, demonstrating improved performance with reduced resource demands.

\textbf{Flexibility in Pruning and Layer Adaptability:} Our task-specific pruning methodology, as depicted in Figure~\ref{fig:1} identifies varying optimal numbers of embedding states for different tasks. This adaptability is absent in HOMODISTIL, which operates under a task-agnostic paradigm. Furthermore, our embedding masks, FFN layer masks enable flexibility in pruning of layers enabling higher pruning in certain layers - an aspect not accommodated by HOMODISTIL, which maintains consistent layer sizes across all layers.

\textbf{Metric Constraints :} Our method has been tested using both parameters and FLOPs constraints, recognizing that each constraint affects computational operations differently. We emphasize that a lower number of layers is often preferred for inference speedup, while hardware limitations, especially in edge devices like Raspberry Pi, typically align with FLOPs constraints. In contrast, HOMODISTIL manages local sparsity within each layer, focusing on controlling total model parameters.

\textbf{Explainability and VIB-based Approach:} Through our VIB-based method, our approach offers explainability by identifying and pruning redundant units such as attention heads. This feature distinguishes our methodology from HOMODISTIL, which does not explicitly provide explainability for its pruning decisions.
\section{Further Ablation Experiments on GLUE dataset with BERT}\label{ap:abl}

\subsection{How many heads are enough without losing accuracy?}
\begin{figure}[h]
\centerline{\includesvg[inkscapelatex=false,width=0.3\textwidth]{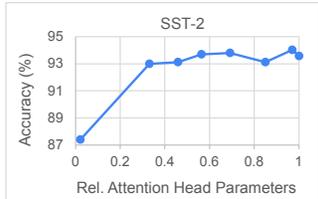}}
    \caption{ When only attention heads are pruned, it is seen that for downstream tasks like SST-2, only 30\% of the total heads are sufficient to retain performance.\label{fig:attheads}}
\end{figure}

On pruning just the attention heads in each layer of the transformer, we see that most GLUE tasks need less than 40\% of the attention heads to retain the performance of the model. As seen in Figure~\ref{fig:attheads}, for task SST-2, 30\% of the heads seem to be enough to retain the performance of the large teacher model.

\subsection{Ablating Embedding states pruning}
We compare the performance of our pruning method specifically with that of CoFi to gauge the efficiency of our VIB-based pruning method with flexible layer sizes in Table~\ref{tab:ablEmb}. For fairness, we ablate the Embedding states pruning and just prune the other modules of the transformer. The teacher which is also the initial student being pruned is the kept same for both the methods.
\begin{table}[h] %[20]{R}{0.4\textwidth}
\centering
\resizebox{0.35\textwidth}{!}{
\begin{tabular}{lcc}
\hline
Tasks & CoFi & Our Method \\
%&&(excluding embedding pruning)\\
\hline
SST-2 &90.6 &91.2 \\
MRPC &82.6 & 82.8\\
RTE &64.7 & 65.3\\
QNLI &86.1 &86.9 \\
\hline
\end{tabular}}
\caption{Results showing efficiency of VIB-based pruning coupled with flexible layer sizes in our method. Both methods prune the same structural components- all components except Embedding parameters of BERT base. All models have about 5 million parameters (excluding embeddings)  \label{tab:ablEmb}}
\end{table}

\subsection{Analysis of total prune time}
The total compression time mentioned in Table~\ref{tab:glue} includes the pruning and finetuning time. The breakdown of the same is given in Table~\ref{tab:time}.
\begin{table}[h] %[12]{L}{0.3\textwidth}
\centering
\resizebox{0.35\textwidth}{!}{
\begin{tabular}{l|cc}
\hline
 & Prune & Fine-tune \\
\hline
GLUE-high& 23\%  & 78\%  \\
SQuAD & 20\% & 80\%   \\
\hline
\end{tabular}}
\caption{Breakdown of prune time and finetuning time on pruning BERT-base \label{tab:time}}
\end{table}

\section{Further Qualitative Analysis}\label{ap:qual}
\subsection{Pruning pattern}\label{sec:prune}
\begin{figure*}[h]
\centerline{\includesvg[inkscapelatex=false,width=1.0\textwidth]{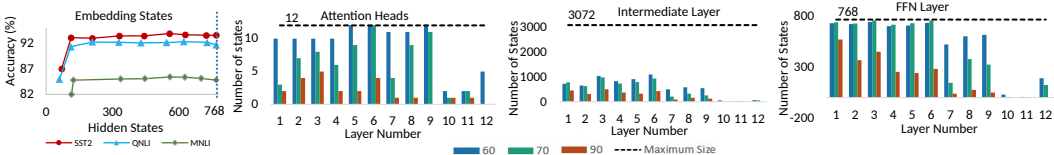}}
    \caption{The average pruning pattern across different structural elements of BERT for GLUE tasks. Left plot: Shows pruned models (dots) retain performance at more than 80\% lower embedding states.}
    \label{fig_pattern}
    % \vspace{-4mm}
\end{figure*}

%our approach allows flexible sizes; better representation
We analyze the pruned model structure obtained with our primary compression method.
Figure~\ref{fig_pattern} shows the remaining attention heads, intermediate dimensions and hidden states in FFN layers at three different sparsity levels(60\%, 70\% and 90\%) averaged on five datasets within GLUE. From the left-most plot in Figure~\ref{fig_pattern}, we note that the over-parameterized embedding layer can be compressed such that 20\% of the total embedding states are enough to retain performance of the un-pruned model. We also observe from the other plots in the figure that there is a significant decrease in the intermediate dimensions at all sparsity levels. At higher sparsity levels (at 90\%), the latter intermediate and FFN layers are deemed redundant and eliminated.
Most of the pruning happens from the latter layers. 
\subsection{More examples of redundant heads} \label{sec:redundant_heads}
Figure~\ref{fig_tokens} shows more examples of redundant heads with a different input sentence than shown in the main paper. Repetitive heads with attention to CLS tokens, attention to own tokens and broad attention in deemed redundant and can be pruned out from the teacher model. 
\subsection{Shift of attention to next or previous token}\label{ap:att_tok}
\begin{figure}[h]
\centerline{\includesvg[inkscapelatex=false,width=\textwidth]{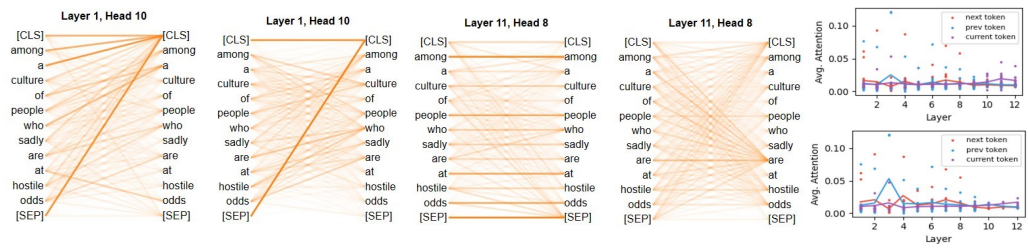}}
    \caption{Another example of heads that are deemed redundant and pruned. Repetitive Heads with tokens attending to the CLS token, tokens attending to themselves and heads with broad attention are deemed redundant. Plots show the average attention paid to next, current and previous token in (Top:) Un-pruned model and (Bottom:) pruned model}
    \label{fig_tokens}
    % \vspace{-6mm}
\end{figure}

Figure \ref{fig_tokens} shows how average attention of input tokens in the compressed models change from current token to next or previous token. Attending to current token does not give the model any new information while focusing more on previous or next token might yield better performance.
\subsection{Divergent Behavior Among Attention Heads.}\label{ap:dis}
\begin{figure}[h]
\centerline{\includesvg[inkscapelatex=false,width=\columnwidth]{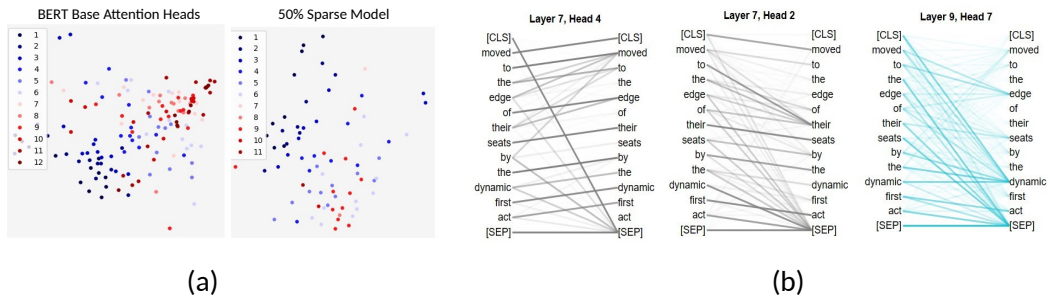}}
    \caption{(a) Comparison between full BERT-base and 50\% pruned model attention heads in 2-D space. Distance between points reflect average Jensen-Shannon divergences between head representations. (b) Heads in pruned model exhibit sparser, divergent behavior compared to un-pruned model. Heads retained in the 50\% pruned model showing attention among input tokens from SST-2 dataset.}
    \label{fig_heads}
  \end{figure}

We calculate the Jensen-Shannon Divergence $ \sum_{\text {token } \in \text { data }} J S\left(\mathrm{H}_i(\text {token}), \mathrm{H}_j(\text {token})\right)$ to measure the pairwise distances between attention head distributions over input data. Using multidimensional scaling~\cite{kruskal1964multidimensional}, we create a two-dimensional representation, ensuring the resulting embeddings closely approximate the Jensen-Shannon distances between the corresponding heads. We then visualize the behavior among heads in Figure~\ref{fig_heads}. In the un-pruned BERT model clustering in layers show similar head behavior. Layers 9 to 12 and layers 1 to 4 exhibit close grouping among themselves suggesting analogous behavior in adjacent layers.
Conversely, in the pruned model, heads from layers 1 to 4 show a more dispersed distribution, indicating a broader range of behaviors. In the later layers (8 to 11), heads appear sparse and widely dispersed with layer 12 being entirely pruned. This validates our approach, demonstrating its effectiveness in removing redundant heads within a layer and entire layers with similar behavior. Moreover, our method aligns with the interpretation by ~\cite{burgess2018understanding} suggesting that VIB encourages the acquisition of more disentangled representations.
\end{document}